# Is the winner really the best?
# A critical analysis of common research practice in biomedical image analysis competitions

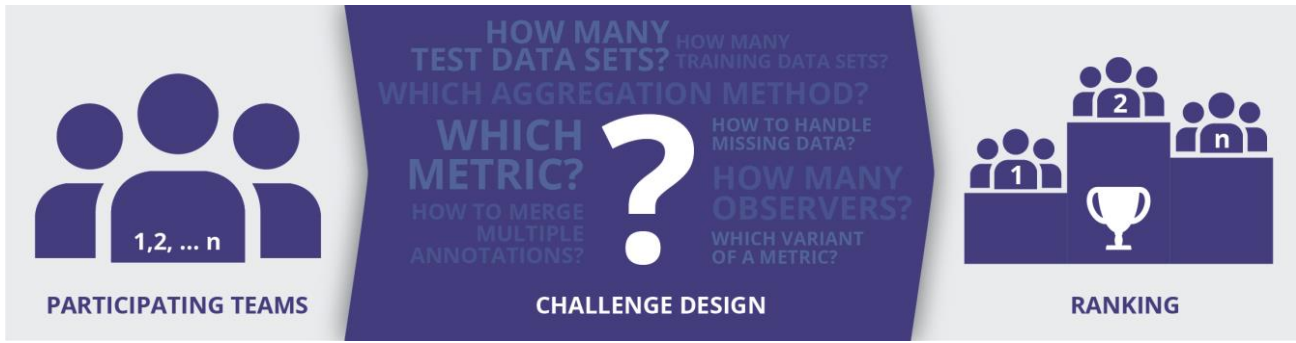

Lena Maier-Hein[1,*,x], Matthias Eisenmann[1,x], Annika Reinke[1], Sinan Onogur[1], Marko Stankovic[1], Patrick Scholz[1], Tal Arbel[2], Hrvoje Bogunovic[3], Andrew P. Bradley[4], Aaron Carass[5], Carolin Feldmann[1], Alejandro F. Frangi[6], Peter M. Full[1], Bram van Ginneken[7], Allan Hanbury[8], Katrin Honauer[9], Michal Kozubek[10], Bennett A. Landman[11], Keno März[1], Oskar Maier[12], Klaus Maier-Hein[13], Bjoern H. Menze[14], Henning Müller[15], Peter F. Neher[13], Wiro Niessen[16], Nasir Rajpoot[17], Gregory C. Sharp[18], Korsuk Sirinukunwattana[19], Stefanie Speidel[20], Christian Stock[21], Danail Stoyanov[22], Abdel Aziz Taha[23], Fons van der Sommen[24], Ching-Wei Wang[25], Marc-André Weber[26], Guoyan Zheng[27], Pierre Jannin[28,x], Annette Kopp-Schneider[29,x]

[1] Division of Computer Assisted Medical Interventions (CAMI), German Cancer Research Center (DKFZ), Heidelberg, Germany

[2] Centre for Intelligent Machines, McGill University, Montreal, QC, Canada

[3] Christian Doppler Laboratory for Ophthalmic Image Analysis, Department of Ophthalmology, Medical University Vienna, Vienna, Austria

[4] Science and Engineering Faculty, Queensland University of Technology, Brisbane, Queensland, Australia

[5] Department of Electrical and Computer Engineering, Department of Computer Science, Johns Hopkins University, Baltimore, USA

[6] CISTIB - Centre for Computational Imaging and Simulation Technologies in Biomedicine, The University of Sheffield, Sheffield, United Kingdom

[7] Department of Radiology and Nuclear Medicine, Medical Image Analysis, Radboud University Center, Nijmegen, The Netherlands

[8] Institute of Information Systems Engineering, TU Wien, Vienna, Austria; Complexity Science Hub, Vienna, Austria

[9] Heidelberg Collaboratory for Image Processing (HCI), Heidelberg University, Heidelberg, Germany

[10] Centre for Biomedical Image Analysis, Masaryk University, Brno, Czech Republic

[11] Electrical Engineering, Vanderbilt University, Nashville, Tennessee, USA

[12] Institute of Medical Informatics, Universität zu Lübeck, Lübeck, Germany

[13] Division of Medical Image Computing (MIC), German Cancer Research Center (DKFZ), Heidelberg, Germany

[14] Institute for Advanced Studies, Department of Informatics, Technical University of Munich, Munich, Germany

[15] Information System Institute, HES-SO, Sierre, Switzerland

[16] Departments of Radiology, Nuclear Medicine and Medical Informatics, Erasmus MC, Rotterdam, The Netherlands

[17] Department of Computer Science, University of Warwick, Coventry, United Kingdom

[18] Department of Radiation Oncology, Massachusetts General Hospital, Boston, Massachusetts, USA

[19] Institute of Biomedical Engineering, University of Oxford, Oxford, United Kingdom

[20] Division of Translational Surgical Oncology (TCO), National Center for Tumor Diseases Dresden, Dresden, Germany

[21] Division of Clinical Epidemiology and Aging Research, German Cancer Research Center (DKFZ), Heidelberg, Germany

[22] Centre for Medical Image Computing (CMIC) & Department of Computer Science, University College London, London, UK

[23] Data Science Studio, Research Studios Austria FG, Vienna, Austria

[24] Department of Electrical Engineering, Eindhoven University of Technology, Eindhoven, The Netherlands

[25] AIExplore, NTUST Center of Computer Vision and Medical Imaging, Graduate Institute of Biomedical Engineering, National Taiwan University of Science and Technology, Taipei, Taiwan

[26] Institute of Diagnostic and Interventional Radiology, University Medical Center Rostock, Rostock, Germany

[27] Institute for Surgical Technology and Biomechanics, University of Bern, Bern, Switzerland

[28] Univ Rennes, Inserm, LTSI (Laboratoire Traitement du Signal et de l'Image) - UMR_S 1099, Rennes, France

[29] Division of Biostatistics, German Cancer Research Center (DKFZ), Heidelberg, Germany

[*] Please send correspondence to: l.maier-hein@dkfz.de

[x] Shared first/senior authors

*Abstract: International challenges have become the standard for validation of biomedical image analysis methods. Given their scientific impact, it is surprising that a critical analysis of common practices related to the organization of challenges has not yet been performed. In this paper, we present a comprehensive analysis of biomedical image analysis challenges conducted up to now. We demonstrate the importance of challenges and show that the lack of quality control has critical consequences. First, reproducibility and interpretation of the results is often hampered as only a fraction of relevant information is typically provided. Second, the rank of an algorithm is generally not robust to a number of variables such as the test data used for validation, the ranking scheme applied and the observers that make the reference annotations. To overcome these problems, we recommend best practice guidelines and define open research questions to be addressed in the future.*

# 1. Introduction

Biomedical image analysis has become a major research field in biomedical research, with thousands of papers published on various image analysis topics including segmentation, registration, visualization, quantification, object tracking and detection [Ayache and Duncan 2016, Chen et al. 2017]. For a long time, validation and evaluation of new methods were based on the authors' personal data sets, rendering fair and direct comparison of the solutions impossible [Price 1986]. The first known efforts to address this problem date back to the late 90s [West et al. 1997], when Jay West, J Michael Fitzpatrick and colleagues performed an international comparative evaluation on intermodality brain image registration techniques. To ensure a fair comparison of the algorithms, the participants of the study had no knowledge of the gold standard results until after their results had been submitted. A few years later, the ImageCLEF[1] [Müller et al. 2004] evaluation campaign introduced a challenge on medical image retrieval [Kalpathy-Cramer et al. 2015], based on experiences in the text retrieval domain where systematic evaluation had been performed since the 1960s [Cleverdon 1960]. About one decade ago, a broader interest in biomedical challenge organization arose with the first grand challenge that was organized in the scope of the international conference on Medical Image Computing and Computer Assisted Intervention (MICCAI) 2007 [Heimann et al. 2009]. Over time, research practice began to change, and the number of challenges organized annually has been increasing steadily (Fig. 1a)), with currently about 28 biomedical image analysis challenges with a mean of 4 tasks conducted annually. Today, biomedical image analysis challenges are often published in prestigious journals (e.g. [Chenouard et al. 2014, Sage et al. 2015, Menze et al. 2015, Ulman et al. 2017, Maier-Hein et al. 2017]) and receive a huge amount of attention with hundreds of citations and thousands of views. Awarding the winner with a significant amount of prize money (up to €1 million on platforms like Kaggle [kaggle]) is also becoming increasingly common.

This development was a great step forward, yet the increasing scientific impact [Tassey et al. 2010, Tsikrika, Herrera and Müller 2011] of challenges now puts huge responsibility on the shoulders of the challenge hosts that take care of the organization and design of such competitions. The performance of an algorithm on challenge data is essential, not only for the acceptance of a paper and its impact on the community, but also for the individuals' scientific careers, and the potential that algorithms can be translated into clinical practice. Given that this is so important, it is surprising that no commonly respected quality control processes for biomedical challenge design exist to date. Similar problems exist in other research communities, such as computer vision and machine learning.

In this paper, we present the first comprehensive evaluation of biomedical image analysis challenges based on 150 challenges conducted up until the end of 2016. It demonstrates the crucial nature of challenges for the field of biomedical image analysis, but also reveals major problems to be addressed: Reproduction, adequate interpretation and cross-comparison of results are not possible in the majority of challenges, as only a fraction of the relevant information is reported and challenge design (e.g. a choice of metrics and methods for rank computation) is highly heterogeneous. Furthermore, the rank of an algorithm in a challenge is sensitive to a number of design choices, including the test data sets used for validation, the observer(s) who annotated the data and the metrics chosen for performance assessment, as well as the methods used for aggregating values.

---
[1] http://www.imageclef.org/

## 2. Results

### 150 biomedical image analysis challenges

Up until the end of 2016, 150 biomedical image analysis challenges that met our inclusion criteria (see *Online Methods)* were conducted with a total of 549 different image analysis tasks (see Fig. 1). 56% of these challenges published their results in journals or conference proceedings. The information used in this paper from the remaining challenges was acquired from websites. Most tasks were related to segmentation (70%) and classification (10%) and were organized within the context of the MICCAI conference (50%), and the *IEEE International Symposium on Biomedical Imaging* (ISBI) (following at 34%). The majority of the tasks dealt with 3D (including 3D+t) data (84%), and the most commonly applied imaging techniques were magnetic resonance imaging (MRI) (62%), computed tomography (CT) (40%) and microscopy (12%). The percentage of tasks that used *in vivo*, *in silico*, *ex vivo*, *in vitro*, *post mortem* and *phantom* data was 85%, 4%, 3%, 2%, 2% and 1%, respectively (9%: N/A; 3%: combination of multiple types). The *in vivo* data was acquired from patients in clinical routine (60%), from patients under controlled conditions (9%), from animals (8%), from healthy human subjects (5%), or from humans under unknown (i.e. not reported) conditions (32%). While training data is typically provided by the challenge organizers (85% of all tasks), the number of training cases varies significantly across the tasks (median: 15; interquartile range (IQR): (7, 30); min: 1, max: 32,468). As with the training cases, the number of test cases varies across the tasks (median: 20; IQR: (12, 33); min: 1, max: 30,804). The median ratio of training cases to test cases was 0.75. The test data used differs considerably from the training data, not only in quantity but also in quality. For 73% of all tasks with human or hybrid reference generation, multiple observers have annotated the reference data. In these cases, an image was annotated by a median of 3 (IQR: (3, 4), max: 9) observers.

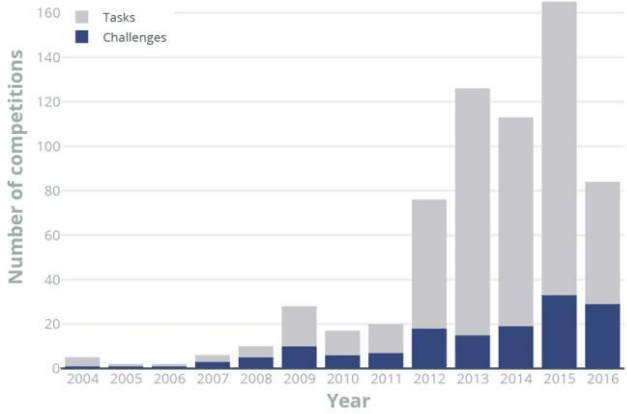
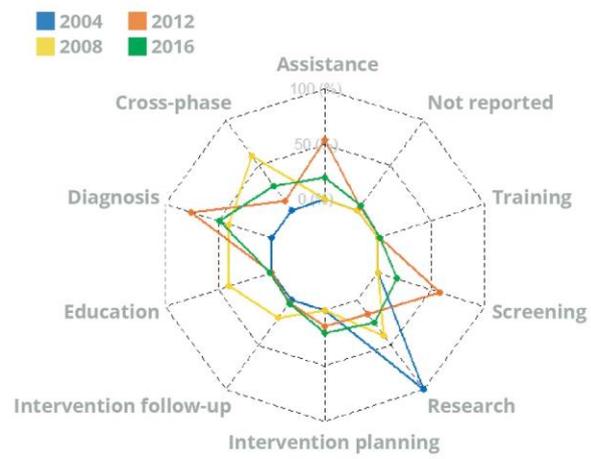
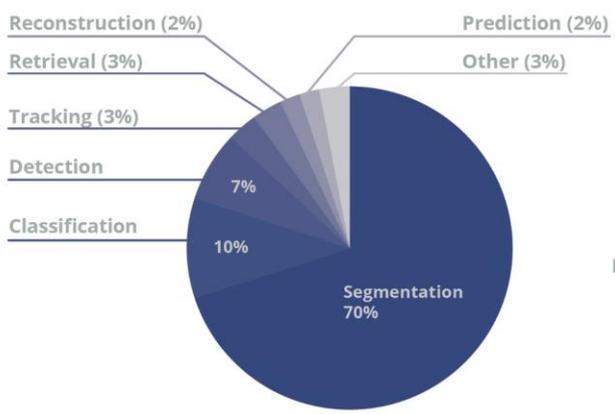
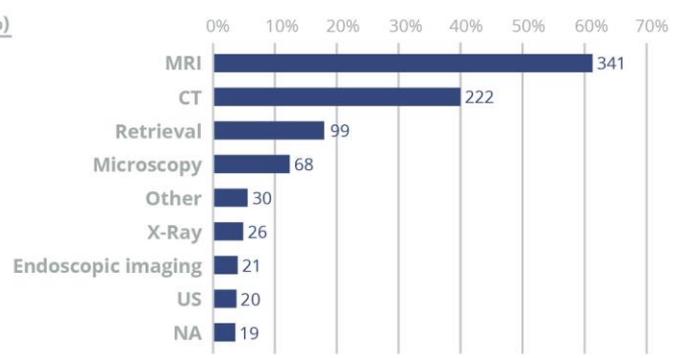
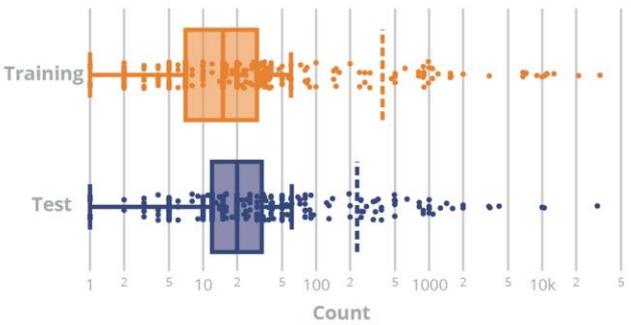
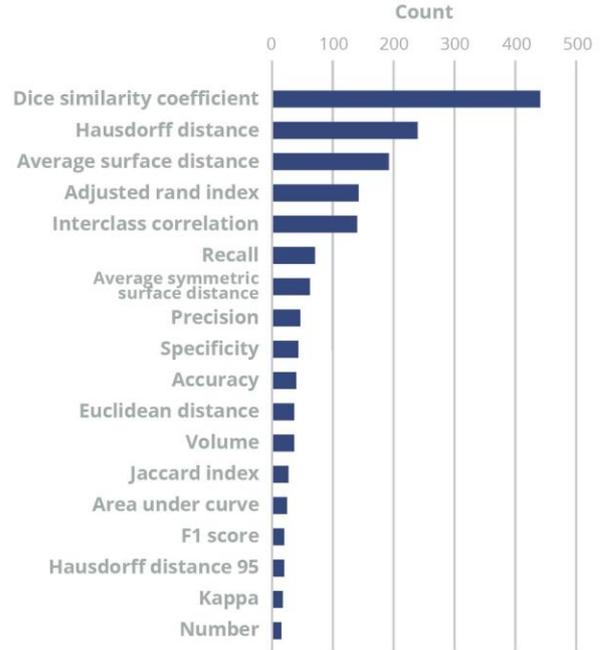
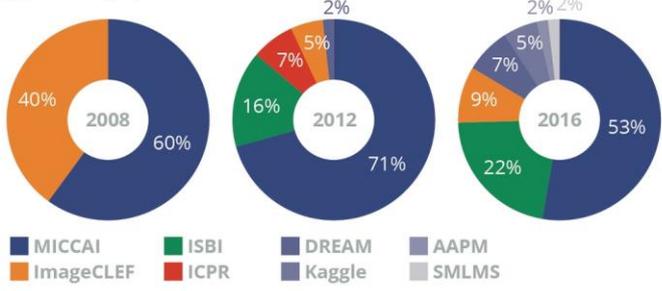

*Fig. 1: Overview of biomedical image analysis challenges. (a) Number of competitions (challenges and tasks) organized per year, (b) fields of application, (c) algorithm categories assessed in the challenges, (d) imaging techniques applied, (e) number of training and test cases used, (f) most commonly applied metrics for performance assessment used in at least 5 tasks and (g) platforms (e.g. conferences) used to organize the challenges for the years 2008, 2012 and 2016.*

# Half of the relevant information is not reported

We identified the relevant parameters that characterize a biomedical challenge following an ontological approach (see *Online Methods*). This yielded a total of 53 parameters corresponding to the categories *challenge organization*, *participation conditions*, *validation objective, study conditions, validation data sets, assessment method* and *challenge outcome* (see Tab. 1). A biomedical challenge task reported a median of 64% (IQR: (49%, 72%); min: 21%, max: 92%) of these parameters. 6% of the parameters were reported for all tasks and 43% of all parameters were reported for less than 50% of all tasks. The list of parameters which are generally not reported includes some that are crucial for interpretation of results. For example, 8% of all tasks providing an aggregated ranking across multiple metrics did not report the rank aggregation method they used (i.e. the method according to which the winner has been determined). 85% of the tasks did not give instructions on whether training data provided by challenge organizers may have been supplemented by other publicly available or private data, although the training data used is key to the success of any machine learning algorithm (see e.g. [Russakovsky et al. 2015]). In 66% of all tasks, there was no description on how the reference (i.e. gold standard) annotation was performed although the quality of annotation in the field of biomedical image analysis varies dependent on the user [Grünberg et al. 2017]. 45% of tasks with multiple annotators did not describe how the annotations were aggregated. Also, the level of expertise of the observers that annotated the reference data was often (19%) not described.

| Category | Parameter name | Coverage [%] |
|---|---|---|
| **Challenge organization** | Challenge name* | 100 |
| | Challenge website* | 99 |
| | Organizing institutions and contact person* | 97 |
| | Life cycle type* | 100 |
| | Challenge venue or platform | 99 |
| | Challenge schedule* | 81 |
| | Ethical approval* | 32 |
| | Data usage agreement | 60 |
| **Participation conditions** | Interaction level policy* | 62 |
| | Organizer participation policy* | 6 |
| | Training data policy* | 15 |
| | Pre-evaluation method | 5 |
| | Submission format* | 91 |
| | Submission instructions | 91 |
| | Evaluation software | 26 |
| **Validation objective** | Field(s) of application* | 97 |
| | Task category(ies)* | 100 |
| | Target cohort* | 65 |
| | Algorithm target(s)* | 99 |
| | Data origin* | 98 |
| | Assessment aim(s)* | 38 |

| Study conditions | Validation cohort* | 88 |
|---|---|---|
| | Center(s)* | 44 |
| | Imaging modality(ies)* | 99 |
| | Context information* | 35 |
| | Acquisition device(s) | 25 |
| | Acquisition protocol(s) | 72 |
| | Operator(s) | 7 |
| Validation data sets | Distribution of training and test cases* | 18 |
| | Category of training data generation method* | 89 |
| | Number of training cases* | 89 |
| | Characteristics of training cases* | 79 |
| | Annotation policy for training cases* | 34 |
| | Annotator(s) of training cases* | 81 |
| | Annotation aggregation method(s) for training cases* | 29 |
| | Category of reference generation method* | 87 |
| | Number of test cases* | 77 |
| | Characteristics of test cases* | 77 |
| | Annotation policy for test cases* | 34 |
| | Annotator(s) of test cases* | 78 |
| | Annotation aggregation method(s) for test cases* | 34 |
| | Data pre-processing method(s) | 24 |
| | Potential sources of reference errors | 28 |
| Assessment method | Metric(s)* | 96 |
| | Justification of metrics* | 23 |
| | Rank computation method* | 36 |
| | Interaction level handling* | 44 |
| | Missing data handling* | 18 |
| | Uncertainty handling* | 5 |
| | Statistical test(s)* | 6 |
| Challenge outcome | Information on participants | 88 |
| | Results | 87 |
| | Publication | 74 |

*Tab. 1: List of parameters that were identified as relevant when reporting a challenge along with the percentage of challenge tasks for which information on the parameter has been reported (red: < 50%; orange: between 50% and 90%; green: > 90%). Parameter definitions can be found in Suppl.1.*
*\*: Parameters used for structured challenge submission for the MICCAI 2018 challenges.*

## Large variability in challenge design

In total, 97 different metrics have been used for performance assessment (three on average per task). Metric design is very heterogeneous, particularly across comparable challenges, and justification for a particular metric is typically (77%) not provided. Roughly half of all metrics (51%) were only applied on a single task. Even in the main field of medical image segmentation, 33% of the 38 different metrics used were only applied once. The fact that different names may sometimes refer to the same metric was compensated for in these computations. 38% of all tasks provided a final ranking of the participants and thus determined a challenge winner. 57% of all tasks that provide a ranking do so on the basis of a single metric. In this case, either metric-based (aggregate, then rank; 70%) or case-based (rank per case, then aggregate; 1%) is typically performed (see *Online Methods*). Overall, 10 different methods for determining the final rank (last step in computation) of an algorithm based on multiple metrics were applied.

## Minor changes in metrics may make the last the first

Besides the Dice Similarity Coefficient (DSC) [Dice 1945], which was used in 92% of all 383 segmentation tasks (2015: 100%), the Hausdorff Distance (HD) [Huttenlocher et al. 1993, Dubuisson and Jain 1994] is the most commonly applied metric in segmentation tasks (47%). It was used either in its original formulation (42%) or as the 95% variant (HD95) (5%) (38%/8% in the 2015 segmentation challenges). We determined a single-metric ranking based on both versions for all 2015 segmentation challenges and found radical differences in the rankings as shown in Fig. 2a). In one case, the worst-performing algorithm according to the HD (10th place) was ranked first in a ranking based on the HD95.

## Different aggregation methods produce different winners

One central result of most challenges is the final ranking they produce. Winners are considered "state of the art" and novel contributions are then benchmarked according to them. The significant design choices related to the ranking scheme based on one or multiple metric(s) are as follows: whether to perform metric-based (aggregate, then rank) or case-based (rank, then aggregate) and whether to take the mean or the median. Statistical analysis with Kendall's tau (rank correlation coefficient [Kendall 1938]) using all segmentation challenges conducted in 2015 revealed that the test case aggregation method has a substantial effect on the final ranking, as shown in Fig. 2b)/c). In some cases, almost all teams change their ranking position when the aggregation method is changed. According to bootstrapping experiments (Fig. 3 and 4), single-metric rankings are statistically highly significantly more robust when (1) the mean rather than the median is used for aggregation and (2) the ranking is performed after the aggregation.

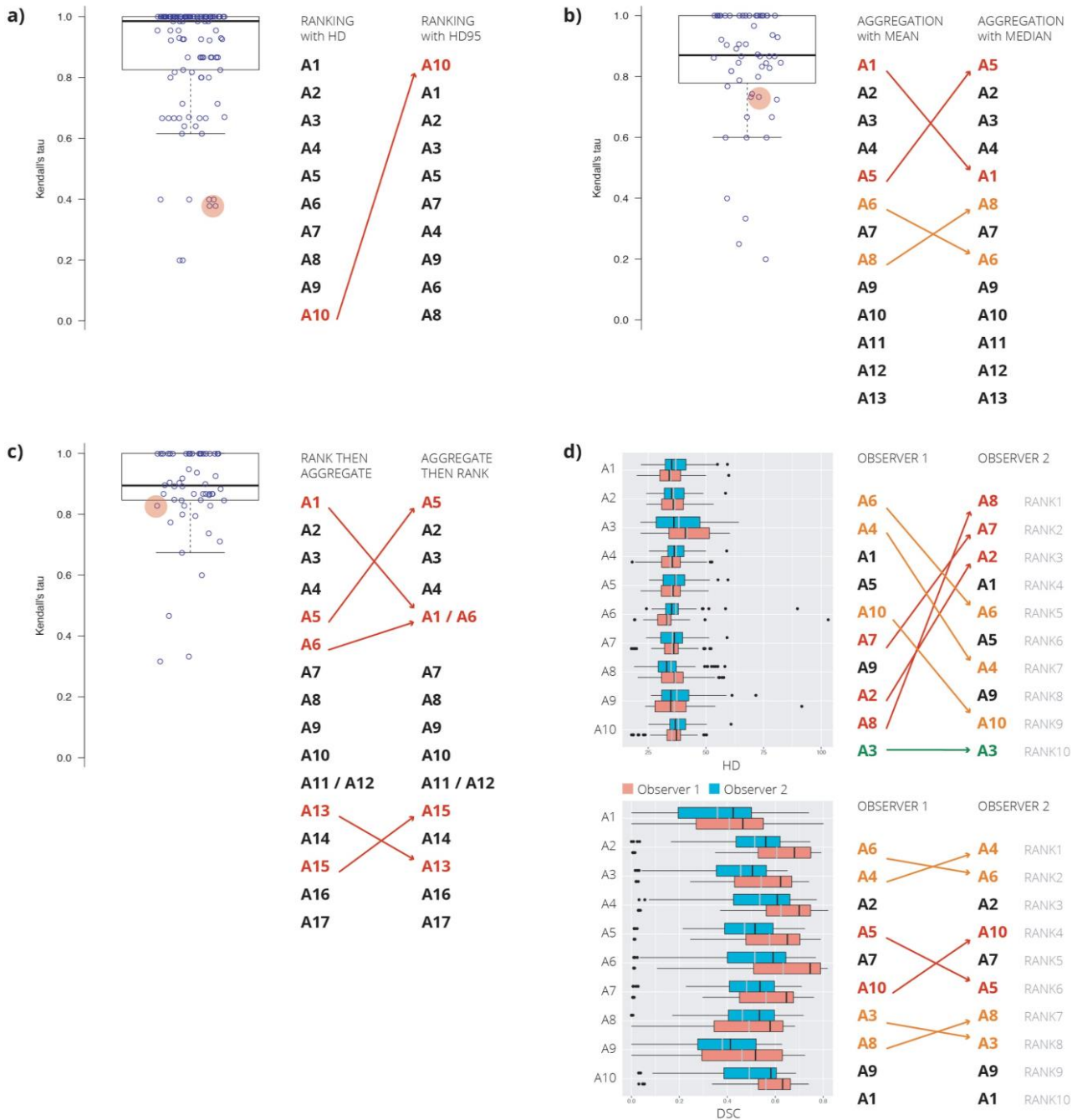

*Fig. 2: Robustness of rankings with respect to several challenge design choices. One data point corresponds to one segmentation task organized in 2015 (n = 56). a) Ranking (metric-based) with the standard Hausdorff Distance (HD) vs. its 95% variant (HD95). b) Mean vs. median in metric-based ranking based on the HD. c) Case-based (rank per case, then aggregate with mean) vs. metric-based (aggregate with mean, then rank) ranking in single-metric ranking based on the HD. d) Metric values per algorithm and rankings for reference annotations performed by two different observers. In the box plots a)-c), descriptive statistics for Kendall's tau, which quantifies differences between rankings (1: identical ranking; -1: inverse ranking), is shown. Key examples illustrate that slight changes in challenge design may lead to the worst algorithm becoming the winner (a) or to almost all teams changing their ranking position (d). Even for relatively high values of Kendall's tau ((b): tau = 0.74; (c): tau = 0.85), critical changes in the ranking may occur.*

## Different annotators produce different winners

In most segmentation tasks (62%), it remains unclear how many observers annotated the reference data. Statistical analysis of the 2015 segmentation challenges, however, revealed that different observers may produce substantially different rankings, as illustrated in Fig. 2(d). In experiments performed with all 2015 segmentation challenges that had used multiple observers for annotation (three tasks with two observers, one with five observers), different observers produced different rankings in 15%, 46% and 62% of the 13 pairwise comparisons between observers, when using a single-metric ranking with mean aggregation based on the DSC, the HD and the HD95, respectively. In these cases, the ranges of Kendall's tau were [0.78, 1], [-0.02, 1] and [0.07,1], respectively.

## Removing one test case can change the winner

Ideally, a challenge ranking should reflect the algorithms' performances for a task and thus be independent of the specific data sets used for validation. However, a re-evaluation of all segmentation challenges conducted in 2015 revealed that rankings are highly sensitive to the test data applied (Fig. 4). According to bootstrapping experiments with the most commonly applied segmentation metrics, the first rank is stable (the winner stays the winner) for 21%, 11% and 9% of the tasks when generating a ranking based on the DSC, the HD or the 95% variant of the HD, respectively. For the most frequently used metric (DSC; 100% of all 2015 segmentation challenges), a median of 15% and up to 100% of the other teams were ranked first in at least 1% of the bootstrap partitions. Even when leaving out only a *single* test case (and thus computing the ranking with one test case less), other teams than the winning team were ranked first in up to 16% of the cases. In one task, leaving a single test case out led to 67% of the teams other than the winning team ranking first.

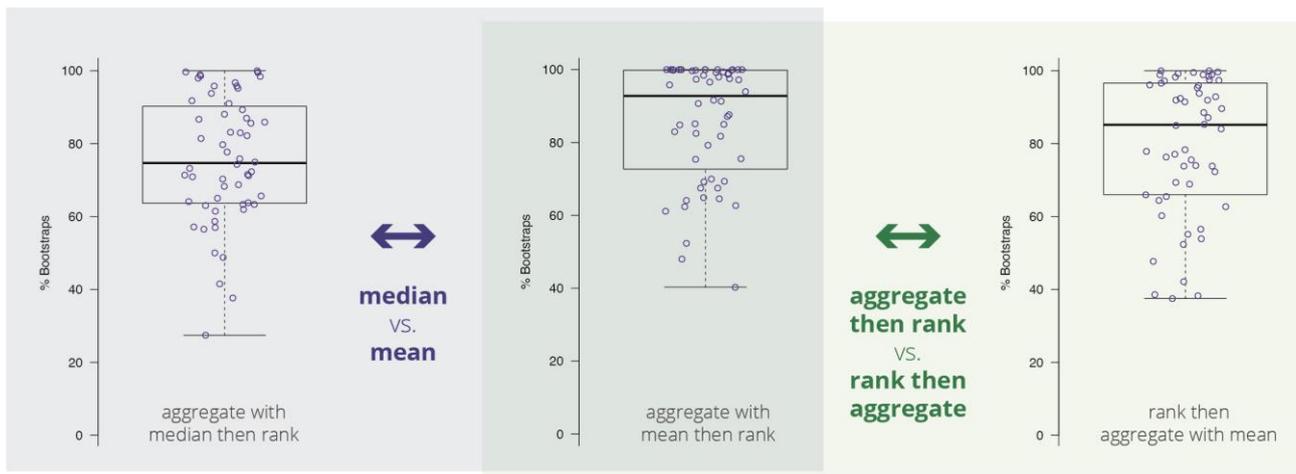

*Fig. 3: The ranking scheme is a deciding factor for the ranking robustness. According to bootstrapping experiments with 2015 segmentation challenge data, single-metric based rankings (those shown here are for the DSC) are significantly more robust when the mean rather than the median is used for aggregation (left) and when the ranking is performed after aggregation rather than before (right). One data point represents the robustness of one task, quantified by the percentage of simulations in bootstrapping experiments in which the winner remains the winner.*

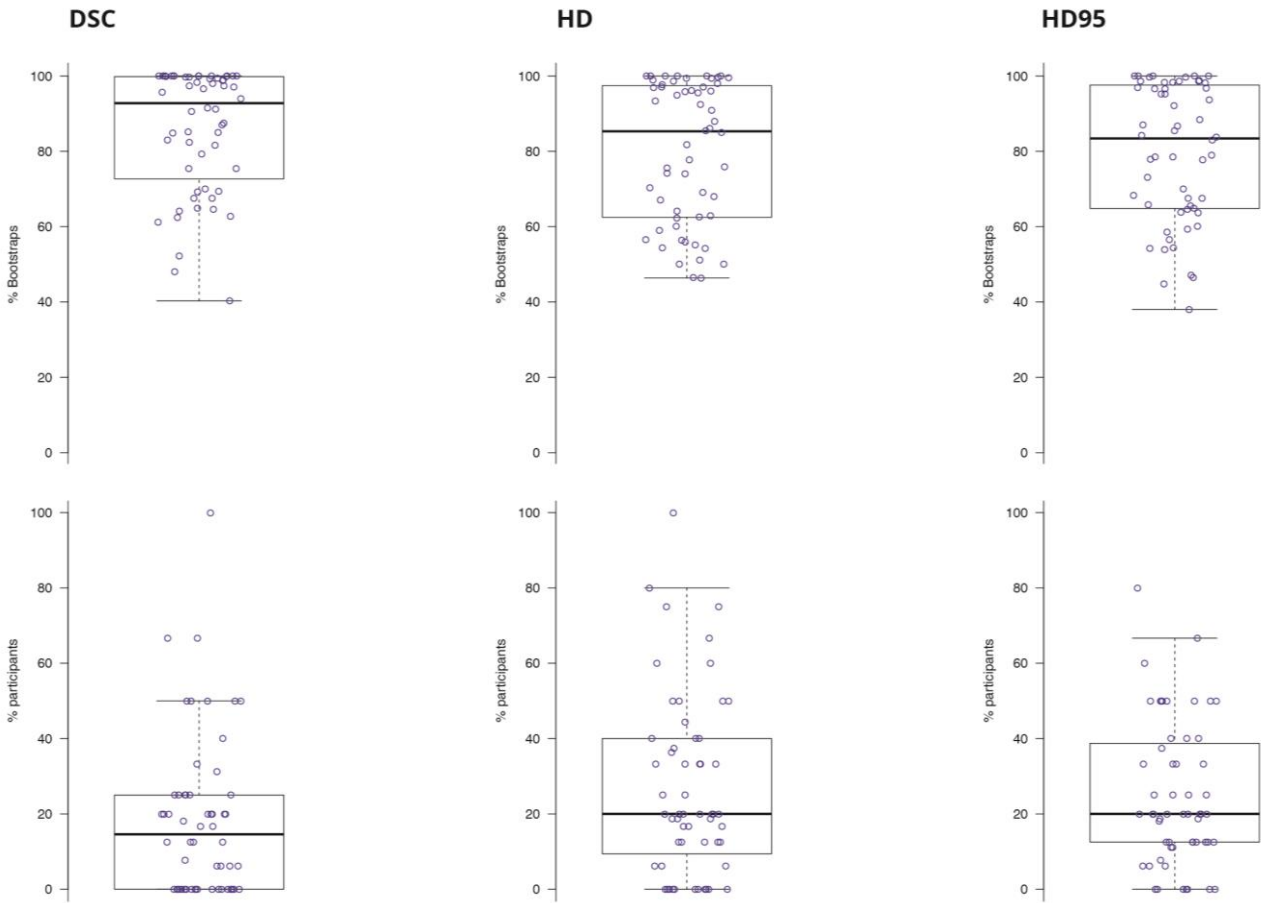

*Fig. 4: Robustness of rankings with respect to the data used when a single-metric ranking scheme based on whether the Dice Similarity Coefficient (DSC) (left), the Hausdorff Distance (HD) (middle) or the 95% variant of the HD (right) is applied. One data point corresponds to one segmentation task organized in 2015 (n = 56). Metric-based aggregation with mean was performed in all experiments. Top: Percentage of simulations in bootstrapping experiments in which the winner (according to the respective metric) remains the winner. Bottom: Percentage of other participating teams that were ranked first in the simulations.*

## Lack of missing data handling allows for rank manipulation

82% of all tasks provide no information about how missing data is handled. While missing data handling is straightforward in case-based aggregation (the algorithms for which no results were submitted receive the last rank for that test case) it is more challenging in metric-based aggregation, especially when no "worst possible value" can be defined for a metric. For this reason, several challenge designs simply ignore missing values when aggregating values. A re-evaluation of all 2015 segmentation challenges revealed that 25% of all 419 non-winning algorithms would have been ranked first if they had systematically just submitted the most plausible results (ranking scheme: aggregate DSC with mean, then rank). In 9% of the 56 tasks, *every single* participating team could have been ranked first if they had not submitted the poorest cases.

## Researchers request quality control

Our experimental analysis of challenges was complemented by a questionnaire (see *Online Methods*). It was submitted by a total of 295 participants from 23 countries. 92% of participants agreed that biomedical challenge design should be improved in general, 87% of all participants would appreciate best practice guidelines, and 71% agreed that challenges should undergo more quality control. A variety of issues were identified for the categories *data*, *annotation*, *evaluation,* and *documentation* (cf. Fig. 5). Many concerns involved the representativeness of the data, the quality of the (annotated) reference data, the choice of metrics and ranking schemes and the lack of completeness and transparency in reporting challenge results. Details are provided in *Suppl. 2* and *3*.

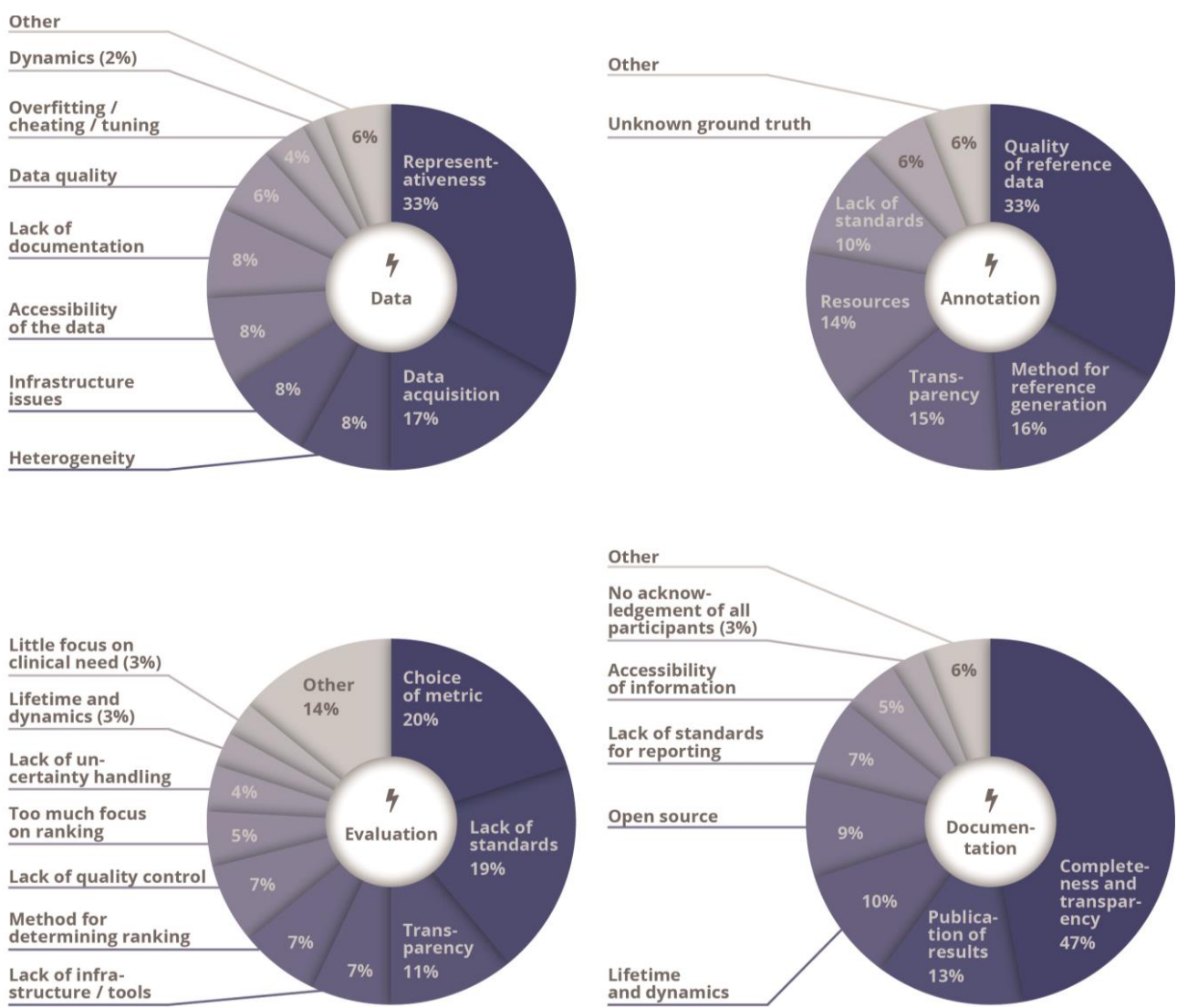

*Fig. 5: Main results of the international questionnaire on biomedical challenges. Issues raised by the participants were related to the challenge data, the data annotation, the evaluation (including choice of metrics and ranking schemes) and the documentation of challenge results.*

## Complete reporting as a first step towards better practices

Based on the findings of this study and the answers to the questionnaire, we have identified several best practice recommendations (see Tab. 2) corresponding to the main problems in biomedical challenge design. The establishment of common standards and clear guidelines is currently hampered by open research questions that still need addressing. However, one primary practice that can be universally recommended is *comprehensive reporting* of the challenge design and results. Our practical and concrete recommendation is therefore to publish the complete challenge design before the challenge by instantiating the list of parameters proposed in this paper (Tab. 1). Three example instantiations are provided in *Suppl. 1*. The MICCAI 2018 satellite event team used the parameter list in the challenge proposal submission system to test its applicability.

The submission system required a potential MICCAI 2018 challenge organizer to instantiate at least 90% of a reduced set of 40 parameters (cf. Tab. 1) that were regarded as essential for judging the quality of a challenge design *proposal*. The median percentage of parameters instantiated was 100% (min: 98%) (16 submitted challenges). Previous tasks analyzed in this paper reported a median of 53% of these parameters.

| Problem | Best practice recommendation(s) | Open research question(s) |
| --- | --- | --- |
| Incomplete reporting | Instantiate the full parameter list (Tab. 1) when reporting on a challenge to maximize transparency, interpretability and reproducibility. | How to describe the data in a structured and standardized manner (e.g. using ontologies)? [Munafò et al. 2017, Zendel et al. 2017, Jannin, Grova and Maurer 2006, Jannin and Korb 2008] |
| Unclear challenge goal | Define a relevant, specific and feasible goal which the challenge will address. Decide on whether to perform an *insight challenge*, the objective of which is to gain insight into a problem and potentially identify a research direction, or a *deployment challenge,* the objective of which is to solve a problem and identify the best-performing algorithms based on a huge benchmarking set. | How to judge the utility (scientific advancement, clinical relevance, biological or clinical insights, implications for patient care, commercial readiness) of a challenge? [Haynes and Haines 1998, Ioannidis 2016, Norman and Verganti 2014, Cutler and McClellan 2001] For *deployment challenges*, how to determine feasibility of clinical deployment in the near, medium and long term? [Shekelle et al. 2006, Black et al. 2011, Garg et al. 2005] |
| Lack of representative-ness | Use data from multiple sources (e.g. sites, devices). Ensure that the selected data collection covers the natural variability of imaged objects. Be aware of the effects of imbalanced training data [Masko and Hensman 2015] when designing the training data set. | How to determine the required number of training/test cases for a given task? [Jain and Chandrasekaran 1982, Raudys 1991, Kalayeh and Landgrebe 2012, Bonnet 2002, Shoukri et al. 2004, Sim and Wright 2005] How to avoid bias in the training/test data? [Deng et al. 2009, Everingham et al. 2010, Chum et al. 2007, Torralba and Efros 2011, Zendel et al. 2017] How to design a challenge that covers the heterogeneity of clinical practice? |
| Low annotation quality | Use multiple annotators per test case. Provide clear guidelines for the annotators. Choose the tools for speeding up annotations carefully as they may lead to bias in the annotations (cf. e.g. [Grünberg et al. 2017]). Find a good compromise between quantity and quality. Consider maximizing annotation quality for the test data ('gold corpus') while increasing quantity at the expense of quality in the training data ('silver corpus'). Assign certified radiologists with standardized training a key role in imaging data annotation to maximize inter-reader agreement [Bamberg et al. 2015, Schlett et al. 2016, Melsaether et al. 2014]. | How to choose the number of observers for a specific task? [Welinder and Perona 2010, Bartlett and Frost 2008, Van den Heuvel and Trip 2002, Walter et al. 1998] How to best combine multiple annotations? [Asman and Landman 2011, Lampert et al. 2016, Sheng et al. 2008, Tian and Zhu 2015, Warfield, Zou and Wells 2004] How to represent, quantify and compensate uncertainty in annotations? [Long et al. 2013, Peng and Zhang 2012] How to provide incentives (especially for clinicians) for data acquisition and annotation? [Von Ahn and Dabbish 2004] How to make data annotation more efficient? [Deng et al. 2009, Everingham et al. 2010, Branson and Perona 2017, Grünberg et al. 2017] |
| Suboptimal metric(s) | Make sure the metrics reflect the challenge goal. Choose metrics that capture the clinically/biologically relevant differences. Be aware of metric-specific biases in favor of/against various properties [Taha and Hanbury 2015]. In segmentation challenges, be aware (1) that the DSC yields more robust rankings than the HD and (2) that the HD yields more robust rankings than the HD95 (Fig. 3). Consider including supplementary *usability* metrics | How to determine the best (variant of a) metric or a set of metrics for a given task? [Taha et al. 2014, Pont-Tuset and Marques 2013, Cehovin et al. 2014] How to better consider clinical relevance in the performance metrics (e.g. by having radiologists quantify the negative effect of segmentation errors)? |
| | related to computation time, memory consumption, number of supported platforms and number of parameters, etc. [Ulman et al. 2017]. | |
| Poor ranking schemes | Be aware of mutually dependent metrics [Cehovin et al. 2014].<br><br>Ensure robust rankings.<br>- Perform metric-based aggregation rather than case-based aggregation to obtain more robust rankings (Fig. 2).<br>- Use the mean rather than the median to obtain more robust rankings (Fig. 2).<br><br>Develop a strategy to handle missing values.<br><br>When applying case-based rankings, consider the tradeoff between robustness (Fig. 2) and good missing value handling.<br><br>Be aware that a statistically significant difference in a metric value may not be clinically/biologically relevant. Vice versa, a clinically relevant difference in performance may not be statistically significant due to small sample size.<br><br>Decide whether a single ranking (one winner) is needed/meaningful.<br><br>Report multiple metric results and provide appropriate visualizations to highlight strengths and weaknesses of different methods [Johannsen et al. 2017, Honauer et al. 2016, Honauer et al. 2015]. | How to handle missing data when aggregating metric values?<br><br>How to determine an appropriate ranking scheme for a given application? [Langville et al. 2012, Pang and Ling 2013, Cabezas et al. 2011, Kristan et al. 2016]<br><br>How to group algorithms (i.e. assign the identical rank) in a sensible manner? [Kristan et al. 2016, Neilson and Yang 2008, Cabezas et al. 2011] |
| Poor uncertainty handling | Quantify the uncertainties of annotations and rankings and make them explicit:<br>- Report inter-observer variability for reference annotations.<br>- Perform bootstrapping to quantify ranking stability (cf. Fig. 3).<br><br>Consider generating fuzzy (probabilistic) reference data and allowing submission of fuzzy results [Taha and Hanbury 2017, Sheng et al. 2008]. | How to incorporate known uncertainties in the reference annotations in the metric computation? [Peng and Zhang 2012]<br><br>How to quantify the uncertainty of a ranking? [Kristan et al. 2016, Pang and Ling 2013, Cabezas et al. 2011] |
| Cheating and overfitting | Publish the challenge design before the challenge according to the parameters in Tab. 1.<br><br>Aim for Docker-based solutions [Hanbury et al. 2017] or on-site challenges to reduce the risk of cheating.<br><br>Otherwise release more test cases than are used for validation (keep the real ones for which annotations are available confidential).<br><br>Do not participate in your own challenge, or otherwise, make the participation transparent.<br><br>Encourage open source release of the algorithms' code. | What is a good lifecycle for a challenge (considering both the dynamics of algorithm development and the overfitting problem)? [Deng et al. 2009, Everingham et al. 2010, Fei-Fei, Fergus and Perona 2006, Perazzi et al. 2016, Dwork et al. 2015] |
| Overhead for participants and organizers | Consider using an existing web-based platform to run the challenge (e.g. [CodaLab], [kaggle], [Grand Challenges], [COVALIC]).<br><br>Consider cloud-based infrastructure for huge data sets and computationally demanding tasks (e.g. [crowdAI]).<br><br>Choose or define a suitable algorithm output format and provide tools for the computation of metrics in this format. Include sample algorithm(s)/workflow(s). | How to establish globally respected standards? |

*Tab. 2: Problems related to current challenge design and organization, best practice recommendations for them and open research challenges (including literature for further reading) relating to them.*

# 3. Discussion

The key findings of this paper can be summarized as follows:

1. Relevance of challenges: Challenges play an increasingly important role in the field of biomedical image analysis, covering a huge range of problems, algorithm classes and imaging modalities (Fig. 1).
2. Challenge reporting: Common practice related to challenge reporting is poor and does not allow for adequate interpretation and reproducibility of results (Tab. 1).
3. Challenge design: Challenge design is very heterogeneous and lacks common standards, although these are requested by the community (Tab. 1, Fig. 5).
4. Robustness of rankings: Rankings are sensitive to a range of challenge design parameters, such as the metric variant applied, the type of test case aggregation performed and the observer annotating the data. The choice of metric and aggregation scheme has a significant influence on the ranking's stability (Fig. 2-4).
5. Best practice recommendations: Based on the findings of this paper and an international survey, we present a list of best practice recommendations and open research challenges. The most universal recommendation is the instantiation of a list of 53 challenge parameters before challenge execution to ensure fairness and transparency along with interpretability and reproducibility of results (Tab. 2).

While this paper concentrates on the field of biomedical image analysis challenges, its impact can be expected to go beyond this field. In fact, many findings of this paper apply not only to challenges but to the topic of validation in general. It may be expected that more effort is typically invested when designing and executing challenges (which, by nature, have a high level of visibility and go hand in hand with publication of the data) compared to the effort invested in performing "in-house studies" dedicated to validation of an individual algorithm. Therefore, concerns involving the meaningfulness of research results in general may be raised. This may also hold true for other research fields, both inside and outside the life sciences, as supported by related literature [Munafò et al. 2017, Ioannidis 2005, Armstrong et al. 2009, Blanco and Zaragoza 2011, Boutros et al. 2014].

In our parameter list (Tab. 1), we have provided a first recommendation to improve the quality of challenge design and its execution and documentation. More international guidelines for challenge organization should be established and a rigorous assessment of challenges organized in the scope of big conferences should be performed in order to move the field towards next-generation biomedical image analysis challenges. To improve data quality, it should be made possible to give open feedback on the data provided by challenges (e.g. ability to report erroneous annotations). Crucially, incentives for organizing and publishing high-quality challenges should be provided. Finally, research dedicated to addressing some of the open issues within challenge design should be actively supported.

In conclusion, challenges are an essential component in the field of biomedical image analysis, but major research challenges and systemic hurdles need to be overcome to fully exploit their potential as valid benchmarking tools.

# Online Methods

**Definitions**

We use the following terms throughout the paper:

Challenge: Open competition on a dedicated scientific problem in the field of biomedical image analysis. A challenge is typically organized by a consortium that issues a dedicated call for participation. A challenge may deal with multiple different tasks for which separate assessment results are provided. For example, a challenge may target the problem of segmentation of human organs in computed tomography (CT) images. It may include several tasks corresponding to the different organs of interest.

Task: Subproblem to be solved in the scope of a challenge for which a dedicated ranking/leaderboard is provided (if any). The assessment method (e.g. metric(s) applied) may vary across different tasks of a challenge.

Case: Dataset for which the algorithm(s) of interest produce one result in either the training phase (if any) or the test phase. It must include one or multiple images of a biomedical imaging modality (e.g. a CT and a magnetic resonance imaging (MRI) image of the same structure) and typically comprises a gold standard annotation (usually required for test cases).

Metric: A measure (not necessarily metric in the strict mathematical sense) used to compute the performance of a given algorithm for a given case, typically based on the known correct answer. Often metrics are normalized to yield values in the interval from 0 (worst performance) to 1 (best performance).

Metric-based vs. case-based aggregation: To rank an algorithm participating in a challenge based on the performance on a set of test cases according to one or multiple metrics, it is necessary to aggregate values to derive a final rank. In single-metric rankings, we distinguish the following two categories, which cover most ranking schemes applied. *Metric-based aggregation* begins with aggregating metric values over all test cases (e.g. with the mean or median). Next, a rank for each algorithm is computed. In contrast, *case-based aggregation* begins with computing a rank for each test case for each algorithm. The final rank is determined by aggregating test case ranks.

**Inclusion criteria**

Inclusion criteria for "Experiment: Comprehensive reporting"

Our aim was to capture all biomedical image analysis challenges that have been conducted up to 2016. We did not include 2017 challenges as our focus is on information provided in scientific papers, which may have a delay of more than a year to be published after challenge execution. To acquire the data, we analyzed the websites hosting/representing biomedical image analysis challenges, namely grand-challenge.org, dreamchallenges.org and kaggle.com as well as websites of main conferences in the field of biomedical image analysis, namely Medical Image Computing and Computer Assisted Intervention (MICCAI), International Symposium on Biomedical Imaging (ISBI), International Society for Optics and Photonics (SPIE) Medical Imaging, Cross Language Evaluation Forum (CLEF), International Conference on Pattern Recognition (ICPR), The American Association of Physicists in Medicine (AAPM), the Single Molecule Localization Microscopy Symposium (SMLMS) and the BioImage Informatics Conference (BII). This yielded a list of 150 challenges with 549 tasks.

Inclusion criteria for "Experiment: Sensitivity of challenge ranking"

All organizers of 2015 segmentation challenges (n = 14) were asked to provide the challenge results (per algorithm and test case) and (re-)compute a defined set of common performance measures, including the Dice Similarity Coefficient (DSC) and the Hausdorff Distance (HD) in the original version [Huttenlocher et al. 1993] and the the 95% variant (HD95) [Dubuisson and Jain 1994]. While the DSC was used in the original design of all 2015 challenges, the HD/HD95 was not always applied. 13 challenges were able to provide the measures as requested. These challenges are composed of 124 tasks in total. The specific inclusion criteria on challenge and task level are provided in Tab. 1 and 2.

| # | Criterion | Number of affected tasks/challenges |
|---|---|---|
| 1 | If a challenge task has on- and off-site part, the results of the part with the most participating algorithms are used. | 1/1 |
| 2 | If multiple reference annotations are provided for a challenge task and *no merged annotation* is available, the results derived from the second[2] annotator are used. | 2/2 |
| 3 | If multiple reference annotations are provided for a challenge task and *a merged annotation* is available, the results derived from the merged annotation are used. | 1/1 |
| 4 | If an algorithm produced invalid values for a metric in all test cases of a challenge task, this algorithm is omitted in the ranking. | 1/1 |

*Tab. 1: Inclusion criteria on challenge level.*

| # | Criterion | Number of excluded tasks |
|---|---|---|
| 1 | Number of algorithms >= 3 | 42 |
| 2 | Number of test cases > 1 (for bootstrapping and cross-validation approaches) | 25 |
| 3 | No explicit argumentation against the usage of Hausdorff Distance as metric | 1 |

*Tab. 2: Inclusion criteria on task level.*

**Challenge parameter list**

One key purpose of this paper was to develop a list of *parameters* that can be instantiated for describing the design and results of a challenge in a comprehensive manner, thus facilitating interpretability and reproducibility of results. To this end, the following procedure was followed:

Initialization: The parameters for describing reference-based validation studies presented in [Jannin, Grova and Maurer 2006] served as an initial set.

---
[2] In one challenge, the first annotator produced radically different annotations from all other observers. This is why we used the second observer of all challenges.

Adding challenge-specific parameters: During analysis of challenge websites and papers, the initial list was complemented such that the information available on a challenge could be comprehensively formalized.

Refinement based on challenge capturing: A tool was designed to formalize existing challenges with the current parameter list. During this process, the list was further refined.

Refinement with international questionnaire: Finally, a questionnaire was designed and sent to all co-authors to finalize the list. All participants were asked to comment on the name, the description, the importance and possible instantiations of each parameter. Adding further parameters was also allowed.

Finalization with ontological modeling: Based on the final list, an ontology for describing biomedical image analysis challenges was developed. The latter was used for structured submission of MICCAI 2018 biomedical challenges.

**Statistical methods**

To quantify the robustness of a ranking, the following statistical methods were used:

Kendall's tau analysis: To quantify the agreement of two rankings (e.g. for two different aggregation methods or two different metric variants), Kendall's tau (also named Kendall's rank correlation or simply tau) [Kendall 1938] was determined as recommended in [Langville et al. 2012]. Tau was designed to be independent of the number of entities ranked and may take values between 1 (perfect agreement, i.e. same ranking) and -1 (reverse ranking).

Bootstrapping: For analysis of the variability of a ranking scheme (e.g. as a function of the metric applied), the following bootstrap approach was chosen: For a given task, the original ranking based on all test cases and a given ranking scheme as well as the winning algorithm according to this ranking scheme was determined. In all analyses, 1000 bootstrap samples were drawn from the data sets and the ranking scheme was applied to each bootstrap sample. It should be noted that on average, 63.2% of distinct data sets are retained in a bootstrap sample. For summary of the ranking scheme variability, the frequency of rank 1 in the bootstrap samples for the original winner ("the winner remains the winner") as well as the proportion of algorithms that achieved rank 1 in the bootstraps but were not winning in the original ranking was determined. Competitions with multiple winners according to the original ranking were not included in the analysis (this occured in just one task). For comparison of the stability of different ranking schemes, the same bootstrap samples were evaluated with different ranking schemes and a paired comparison between the proportion of the "winner remaining the winner" was performed by Wilcoxon signed rank test. Results were considered significant for $p < 0.05$.

Leave-one-out: For a given task, the original ranking based on all test cases and a given ranking scheme and the winning algorithm according to this ranking scheme was determined. The number of datasets was reduced by one and the ranking scheme was applied to this subset of datasets. The same summary measures as for the bootstrapping approach were determined.

Note that we did not rely on results of statistical testing approaches to quantify the stability of a given ranking scheme. The reasons for this decision were the following:

(a) The number of datasets varies widely between different tasks and due to correlation of power and sample size, results of statistical tests between different tasks are not comparable by design.

(b) If one was to use statistical testing, the appropriate approach would be to use a mixed model with a random factor for the data set and test the global hypothesis that all algorithms produce the same result, followed by post-hoc all pairwise comparisons. Pairwise comparisons would have to be adjusted for multiplicity and adjustment depends on the number of algorithms in the task. Again, results of statistical testing between different tasks are not comparable by design.

(c) We have evaluated the concordance of the bootstrap analysis for variability of ranking with a statistical testing approach and found examples where there was a highly significant difference between the winner and the second, but bootstrap analysis showed that ranking was very variable, and vice versa.

Boxplots with and without dots were produced to visualize results. In all boxplots, the boldfaced black line represents the median while the box represents the first and third quartile. The upper whisker extends to the largest observation <= Median +1.5 IQR, and likewise the lower whisker to the smallest observation >= Median -1.5 IQR. In horizontal boxplots, the mean is shown in addition as boldfaced grey line.

All statistical analyses were performed with R version 3.4.3 (The R Foundation for Statistical Computing 2017). The figures were produced with Excel, R, Plotly (Python) and Adobe Illustrator 2017.

**Experiment: Comprehensive reporting**

The key research questions corresponding to the comprehensive challenge analysis were:

RQ1: What is the role of challenges for the field of biomedical image analysis (e.g. How many challenges conducted to date? In which fields? For which algorithm categories? Based on which modalities?)

RQ2: What is common practice related to challenge design (e.g. choice of metric(s) and ranking methods, number of training/test images, annotation practice etc.)? Are there common standards?

RQ3: Does common practice related to challenge reporting allow for reproducibility and adequate interpretation of results?

To answer these questions, a tool for instantiating the challenge parameter list introduced in the previous section was used by five engineers and a medical student to formalize all challenges that met our inclusion criteria as follows: (1) Initially, each challenge was independently formalized by two different observers. (2) The formalization results were automatically compared. In ambiguous cases, when the observers could not agree on the instantiation of a parameter - a third observer was consulted, and a decision was made. When refinements to the parameter list were made, the process was repeated for missing values. Based on the formalized challenge data set, a descriptive statistical analysis was performed to characterize common practice related to challenge design and reporting.

**Experiment: Sensitivity of challenge ranking**

The primary research questions corresponding to the experiments on challenge rankings were:

RQ4: How robust are challenge rankings? What is the effect of

(a) the specific test cases used?
(b) the specific metric variant(s) applied?
(c) the rank aggregation method chosen (e.g. aggregation of metric values with the mean vs median)?
(d) the observer who generated the reference annotation?

RQ5: Does the robustness of challenge rankings vary with different (commonly applied) metrics and ranking schemes?

RQ6: Can common practice on missing data handling be exploited to manipulate rankings?

As published data on challenges typically do not include metric results for individual data sets, we addressed these open research questions by approaching all organizers of segmentation challenges conducted in 2015 and asking them to provide detailed performance data on their tasks (124 in total). Note in this context that *segmentation* is by far the most important algorithm category (70% of all biomedical image analysis challenges) as detailed in the *Results* section. Our comprehensive challenge analysis further revealed *single-metric ranking* with *mean* and *metric-based aggregation* as the most frequently used ranking scheme. This is hence considered the default ranking scheme in this paper.

Our analysis further identified the DSC (92%) and the HD (47%) as the most commonly used segmentation metrics. The latter can either be applied in the original version (42%) or the 95% variant (HD95) (5%).

To be able to investigate the sensitivity of rankings with respect to several challenge design choices, the 2015 segmentation challenge organizers were asked to provide the assessment data (results for DSC, HD and HD95) on a per data set basis for their challenge. The research questions RQ4-6 were then addressed with the following experiments:

RQ4: For all 56 segmentation tasks that met our inclusion criteria, we generated single-metric rankings with the default ranking scheme based on the DSC and the HD. We then used Kendall's tau to investigate the effect of changing (1) the metric variant (HD vs HD95), (2) the aggregation operator (mean vs median), (3) the aggregation category (metric-based vs case-based) and (4) the observer (in case multiple annotations were available). Note in this context that we focused on single-metric rankings in order to perform a statistical analysis that enables a valid comparison across challenges.

RQ5: To quantify the robustness of rankings as a function of the metric, we generated single-metric rankings with the default ranking scheme based on the DSC, the HD and the HD95. We then applied bootstrapping and leave-one-out analysis to quantify ranking robustness as detailed in *Statistical Methods.* Analogously, we compared the robustness of rankings for different aggregation methods (metric-based vs case-based) and aggregation operators (mean vs median).

RQ6: 82% of all biomedical image analysis tasks (see *Results*) do not report any information on missing values when determining a challenge ranking. In metric-based ranking (although not reported), it is common to simply ignore missing values. To investigate whether this common practice may be exploited by challenge participants to manipulate rankings, we performed the following analysis: For each algorithm and each task of each 2015 segmentation challenge that met our inclusion criteria, we determined the default ranking and artificially removed those test set results whose DSC was below a threshold of $t = 0.5$. Note that it can be assumed that these cases could have been relatively easily identified by visual inspection without comparing them to the reference annotations. We then compared the new ranking position of the algorithm with the position in the original (default) ranking.

**International Survey**

As a basis for deriving best practice recommendations related to challenge design and organization, we designed a questionnaire (see *Suppl. 2*) to gather known potential issues. It was distributed to colleagues of all co-authors, the challenges chairs of the past 3 MICCAI conferences as well as to the following mailing lists: ImageWorld, the mailing lists of the MICCAI society, the international society for computer aided surgery (ISCAS), the UK Euro-BioImaging project and the conferences Medical Image Understanding and Analysis (MIUA) and Bildverarbeitung für die Medizin (BVM). The link to the questionnaire was further published on grand-challenge.org.

# Supplement 1

| # | Parameter name | Description | Representative instantiations |
|---|---|---|---|
| *Challenge organization* | | | |
| 1 | Challenge name* | Full name of the challenge with year. | Example: MICCAI Endoscopic Vision Challenge 2015 |
| 2 | Challenge website* | URL of challenge website (if any). | − URL to challenge website<br>− Private link to website under construction<br>− No website |
| 3 | Organizing institutions and contact person* | Information on the organizing team including contact person and other team members. | Should include:<br>− Contact person with affiliation<br>− Team members with affiliations |
| 4 | Life cycle type* | Submission cycle of the challenge. Not every challenge closes after the submission deadline (one-time event). Sometimes it is possible to submit results after the deadline (open call) or the challenge is repeated with some modifications (repeated event).<br><br>*Example 1 - Brain tumor segmentation: One-time event*<br><br>*Example 2 - Instrument tracking: Open call*<br><br>*Example 3 - Modality classification in biomedical literature: Repeated event (each year; third time)* | − One-time event<br>− Repeated event<br>− Open call |
| 5 | Challenge venue or platform | Event (e.g. conference) or platform that is associated with the challenge.<br><br>*Example 1 - Brain tumor segmentation: DREAM*<br><br>*Example 2 - Instrument tracking: None (online competition)*<br><br>*Example 3 - Modality classification in biomedical literature: ImageCLEF* | − Medical Image Computing and Computer Assisted Intervention (MICCAI)<br>− International Symposium on Biomedical Imaging (ISBI)<br>− Dialogue on Reverse Engineering Assessments and Methods (DREAM)<br>− Image Cross Language Evaluation Forum (ImageCLEF)<br>− International conference on pattern recognition (ICPR)<br>− Kaggle<br>− The International Society for Optical Engineering (SPIE) Medical Imaging<br>− Single Molecule Localization Microscopy Symposium (SMLMS)<br>− American Association of Physicists in Medicine (AAPM)<br>− BioImage Informatics (BII) |
| 6 | Challenge schedule* | Timetable for the challenge which includes the release of training and test cases, the submission dates, possibly associated workshop days, release of results and other important dates. | Should include:<br>− Training data release(s)<br>− Test data release(s)<br>− Submission deadline<br>− Conference day (if any) |
| 7 | Ethics approval* | Information on ethics approval, preferably Institutional Review Board, location, date and number of the ethics approval.<br><br>*Example 1 - Brain tumor segmentation: <URL to ethics approval>* | − No ethics needed (due to in silico validation)<br>− URL to ethics approval document<br>− No ethics required (data downloaded from a public database) |
| | | *Example 2 - Instrument tracking: Reference to ethics of the data source (data for the challenge is publicly available)* | |
| | | *Example 3 - Modality classification in biomedical literature: Not needed as the images are from biomedical journals and publications require internal ethics approval. In PubMed central (open access biomedical literature), all images can be redistributed when citing the source; each image has a Creative Commons license attached to the image.* | |
| 8 | Data usage agreement | Instructions on how the data can be used and distributed by the teams that participate in the challenge and by others.<br><br>*Example 1 - Brain tumor segmentation: The data may only be used for the challenge itself and may not be redistributed.*<br><br>*Example 2 - Instrument tracking: The data can be reused for other purposes but the challenge has to be mentioned in the acknowledgements.*<br><br>*Example 3 - Modality classification in biomedical literature: <URL to data usage agreement>* | − Challenge data must not be redistributed to persons not belonging to the registered team.<br>− Challenge data may be used for all purposes provided that the challenge is referenced.<br>− URL to data usage agreement |
| *Participation conditions* | | | |
| 9 | Interaction level policy* | Allowed user interaction of the algorithms assessed.<br><br>*Example 1 - Brain tumor segmentation: Both automatic and semi-automatic algorithms can participate in the challenge.*<br><br>*Example 2 - Instrument tracking: Only fully automatic algorithms are allowed to participate.*<br><br>*Example 3 - Modality classification in biomedical literature: Only fully automatic algorithms are allowed.* | − Fully interactive<br>− Semi-automatic<br>− Fully automatic |
| 10 | Organizer participation policy* | Participation policy for members of the organizers' institutes.<br><br>*Example 1 - Brain tumor segmentation: Members of the organizers' institutes may participate but they are not eligible for awards.*<br><br>*Example 2 - Instrument tracking: Members of the organizers' institutes may not participate.*<br><br>*Example 3 - Modality classification in biomedical literature: Members of the organizers' institutes may not participate.* | − Members of the organizers' institutes may participate but they are not eligible for awards and they will not be listed in the leaderboard.<br>− Members of the organizers' institutes may not participate. |
| 11 | Training data policy* | Policy on the usage of training data. The data used to train algorithms may, for example, be restricted to the data provided by the challenge or to publicly available data including (open) pre-trained nets.<br><br>*Example 1 - Brain tumor segmentation: The challenge training data may be complemented by other publicly available data.* | − No policy as no training data is required<br>− No additional data allowed<br>− Publicly available data may be added<br>− Private data may be added<br>− Docker container |
| | | *Example 2 - Instrument tracking: Participants may only use the data provided by the challenge for the training of their algorithms.* <br><br> *Example 3 - Modality classification in biomedical literature: Participants may use their own data but they have to indicate and describe the additional data.* | |
| 12 | Pre-evaluation method | Information on the possibility to evaluate the algorithms before the best runs are to be submitted for an official challenge. <br><br> *Example 1 - Brain tumor segmentation: Results on a pre-test set* <br><br> *Example 2 - Instrument tracking: No pre-evaluation* <br><br> *Example 3 - Modality classification in biomedical literature: No pre-evaluation* | – No pre-evaluation <br> – Private results <br> – Public leaderboard (based on pre-testset) <br> – Results on validation dataset |
| 13 | Submission format* | Method that is used for result submission. <br><br> *Example 1 - Brain tumor segmentation: Participants send the algorithm output to the organizers via email.* <br><br> *Example 2 - Instrument tracking: Docker container* <br><br> *Example 3 - Modality classification in biomedical literature: Participants submit a run file that contains their class for each image.* | – Docker container <br> – Cloud <br> – Upload whole code <br> – Upload executable <br> – Send algorithm output to organizers <br> – API <br> – Evaluation Platform |
| 14 | Submission instructions | Instructions on how and when the participants should generate and prepare their submissions and what should be included at each stage. <br><br> *Example 1 - Brain tumor segmentation: On <date>, each team has to submit a 2-5 pages short paper with a description of their algorithm and the results on the training and test datasets as described in [ref]. There is no limit in the number of submissions.* <br><br> *Example 2 - Instrument tracking: <link to URL>* <br><br> *Example 3 - Modality classification in biomedical literature: As described in [ref].* | – No instructions <br> – Format of submissions <br> – Timeline <br> – Number of resubmissions allowed <br> – Number of different submissions (different methods) per participant allowed <br> – Missing results/cases allowance <br> – Source code requirement |
| 15 | Evaluation software | Information on the accessibility of the organizers' evaluation code. <br><br> *Example 1 - Brain tumor segmentation: Software (executable and source code) publicly available from the moment the challenge starts (also after the challenge has ended): <URL>* <br><br> *Example 2 - Instrument tracking: No evaluation software available (Docker concept)* <br><br> *Example 3 - Modality classification in biomedical literature: Software to be used for result submission. Only available for registered participants in the ongoing challenge.* | – Not available <br> – Publicly available: provide URL <br> – Partially available <br> – Available after registration |

| | *Validation objective* | | |
|---|---|---|---|
| 16 | Field(s) of application* | Medical or biological application that the algorithm was designed for.<br><br>*Example 1 - Brain tumor segmentation: Diagnosis*<br><br>*Example 2 - Instrument tracking: Surgery*<br><br>*Example 3 - Modality classification in biomedical literature: Education* | − Training<br>− Intervention planning<br>− Intervention follow-up<br>− Diagnosis<br>− Screening<br>− Assistance (e.g. tracking tasks)<br>− Research (e.g. cell tracking)<br>− Cross-phase<br>− Education<br>− Prognosis<br>− Prevention<br>− Medical data management |
| 17 | Task category(ies)* | Category(ies) of the algorithms assessed.<br><br>*Example 1 - Brain tumor segmentation: Segmentation*<br><br>*Example 2 - Instrument tracking: Localization*<br><br>*Example 3 - Modality classification in biomedical literature: Classification* | − Segmentation<br>− Classification<br>− Tracking<br>− Retrieval<br>− Detection<br>− Localization<br>− Registration<br>− Reconstruction<br>− Modeling<br>− Simulation<br>− Regression<br>− Stitching<br>− Restoration<br>− Prediction<br>− Denoising |
| 18 | Target cohort* | Description of subjects/objects from whom the data would be acquired in the final application.<br><br>Remark: A challenge could be designed around the task of medical instrument tracking in robotic kidney surgery. While the validation (see parameter validation cohort) could be performed ex vivo in a laparoscopic training environment with porcine organs, the final application (i.e. robotic kidney surgery) would be targeted on real patients with certain characteristics defined by inclusion criteria such as restrictions regarding gender or age.<br><br>*Example 1 - Brain tumor segmentation: Patients diagnosed with glioblastoma that got MRI scans for diagnosis including T1-weighted 3D acquisitions, T1-weighted contrast-enhanced (gadolinium contrast) 3D acquisitions and T2-weighted FLAIR 3D acquisitions.*<br><br>*Example 2 - Instrument tracking: Patients undergoing laparoscopic robotic kidney surgery with the da Vinci Si.*<br><br>*Example 3 - Modality classification in biomedical literature: Biomedical journals from PubMed Central (PMC), i.e. the open access literature indexed in Medline.* | − Healthy volunteers that undergo screening<br>− Patients that undergo laparoscopic surgery<br>− Patients that get an abdominal CT<br>− Patients of a particular database<br>− Patients referred for early Barrett's esophagus cancer without visible abnormalities<br>− Patients attending a state-of-the-art cardiac MRI diagnostic centre<br>− Healthy volunteers that are recruited for a certain study<br>− Patients that get chemotherapy<br>− Men with clinical suspicion of having prostate cancer<br>− Standardized cancer cell lines (such as HeLa)<br>− Physicians that use a da Vinci Si for surgical training in an ex vivo setting<br>− OR team (surgeons, nurses, ...) during liver transplantation<br>− Specific journals with an oncology focus (for retrieval tasks) |
| 19 | Algorithm target(s)* | Structure/subject/object/component that the algorithm focuses on.<br><br>*Example 1 - Brain tumor segmentation: Glioblastoma*<br><br>*Example 2 - Instrument tracking: Robotic* | − Glioblastoma<br>− Hepatocellular carcinoma (HCC)<br>− Vessels<br>− Liver<br>− Tool tip<br>− (Any) Tumor |
| | | *instruments*<br>*Example 3 - Modality classification in biomedical literature: Figures showing medical images in the journal* | – Surgeon<br>– Nurse<br>– Specific cell type<br>– Operating room<br>– Specular reflections<br>– Fiber pathway |
| 20 | Data origin* | Region(s)/part(s) of subject(s)/object(s) from which the data would be acquired in the final application.<br><br>*Example 1 - Brain tumor segmentation: Brain*<br><br>*Example 2 - Instrument tracking: Abdomen*<br><br>*Example 3 - Modality classification in biomedical literature: JPEG images that appeared in the journal as defined by the target cohort.* | – Abdomen<br>– Liver<br>– Thorax<br>– Whole body<br>– Whole operating room<br>– Cortical gray matter<br>– Specific journal (for retrieval tasks)<br>– Blood obtained from forearm |
| 21 | Assessment aim(s)* | Property(ies) of the algorithms aimed to be optimized.<br><br>Remark: Ideally, the metrics used in the study assess the properties of the algorithm as defined by the parameter *assessment aim(s)*. For example, an assessment aim could be targeted on the accuracy of segmentation algorithms. Possible metrics to assess the accuracy include the Dice similarity coefficient (DSC) and the Hausdorff distance (HD).<br><br>*Example 1 - Brain tumor segmentation: Accuracy of enhancing tumor/necrosis/edema segmentation*<br><br>*Example 2 - Instrument tracking: Runtime and robustness*<br><br>*Example 3 - Modality classification in biomedical literature: Accuracy* | – Accuracy<br>– Robustness<br>– Reliability<br>– Precision<br>– Sensitivity<br>– Specificity<br>– Consistency<br>– Runtime<br>– Applicability<br>– Feasibility<br>– Complexity<br>– Usability<br>– User satisfaction<br>– Criteria linked to ergonomics<br>– Integration in (clinical) workflow<br>– Hardware requirements |
| ***Study conditions*** | | | |
| 22 | Validation cohort* | Subject(s)/object(s) from whom/which the data was acquired used to validate the algorithm.<br><br>Remark: While a challenge is typically targeted on humans, validation may exclusively involve porcine models or phantoms.<br><br>*Example 1 - Brain tumor segmentation: Patients with glioblastoma (retrospective analysis)*<br><br>*Example 2 - Instrument tracking: Ex vivo porcine organs in a laparoscopic training environment*<br><br>*Example 3 - Modality classification in biomedical literature: PMC journals papers published between 2010 and 2015* | – Specific mouse model<br>– Porcine model<br>– Physical phantom<br>– Patients under controlled conditions<br>– Patients in clinical routine<br>– Porcine liver (in vitro)<br>– In silico data<br>– Healthy volunteers |
| 23 | Center(s)* | Center(s) or institute(s) in which the data was acquired.<br><br>*Example 1 - Brain tumor segmentation: National Center for Tumor Diseases (NCT) Heidelberg*<br><br>*Example 2 - Instrument tracking: As listed on* | – Centers involved in the xy study<br>– University Clinic xy<br>– Centers that are part of the xy consortium |
| | | *website: <URL>*<br><br>*Example 3 - Modality classification in biomedical literature: All centers that are mentioned in the articles where the JPEG images originate from* | |
| 24 | Imaging modality(ies)* | Imaging technique(s) applied for training/test data acquisition.<br><br>*Example 1 - Brain tumor segmentation: MRI*<br><br>*Example 2 - Instrument tracking: White light endoscopy*<br><br>*Example 3 - Modality classification in biomedical literature: Any medical imaging modality from the following set: <list>* | – Magnetic Resonance Imaging (MRI)<br>– Computed Tomography (CT)<br>– Ultrasound (US)<br>– 3D US<br>– Intravascular US (IVUS)<br>– Positron Emission Tomography (PET)<br>– Light Microscopy (LM)<br>– Electron Microscopy (EM)<br>– X-Ray<br>– Optical Coherence Tomography (OCT)<br>– Endomicroscopy (w/ or w/o dye)<br>– SPECT<br>– Video<br>– Fluoroscopy<br>– Thermography |
| 25 | Context information* | Additional information given along with the images. The information may correspond directly to the image data (e.g. tumor volume), to the patient in general (e.g. gender, medical history) or to the acquisition process (e.g. medical device data during endoscopic surgery, calibration data for an image modality).<br><br>*Example 1 - Brain tumor segmentation: Clinical patient data: {age, gender, ...}*<br><br>*Example 2 - Instrument tracking: API data of robot for each frame and CAD models of instruments*<br><br>*Example 3 - Modality classification for retrieval tasks: None* | – No additional information<br>– Genetic information<br>– Age<br>– Gender<br>– Pathology<br>– Clinical diagnoses<br>– Patient number<br>– Medical record<br>– Weight<br>– BMI<br>– Race<br>– Cancer (sub-)type<br>– Cancer/disease stage<br>– Body weight/height<br>– Smoking status<br>– Clinical treatment details<br>– Lab data<br>– Clinical history<br>– OR device data<br>– Free text, such as the radiology report, the operation report or histopathology report |
| 26 | Acquisition device(s) | Device(s) used to acquire the validation data. This includes details on the device(s) used to acquire the imaging data (parameter imaging modality(ies)) as well as information on additional devices used for validation (e.g. tracking system used in a surgical setting):<br><br>*Example 1 - Brain tumor segmentation: 3 T Philips Achieva scanner and GE Signa 1.5 T scanner*<br><br>*Example 2 - Instrument tracking: da Vinci Si endoscope and NDI Aurora electromagnetic tracking system with standard electromagnetic field generator*<br><br>*Example 3 - Modality classification in biomedical literature: Large variety of devices provided as a list* | – Philips Gyro Scan NT 1.5 Tesla scanner<br>– GE Discovery ST multislice PET/CT scanner<br>– G-EYE Videocolonoscope<br>– MINDRAY DC-30 US scanner<br>– NDI Aurora electromagnetic tracking system with Tabletop field generator<br>– None (e.g. in case of in silico validation)<br>– Unknown (e.g. in case of retrieval tasks) |
| 27 | Acquisition protocol(s) | Relevant information on the imaging process/ data acquisition for each acquisition device. | – Dimension (e.g. 2D, 3D+t)<br>– Timepoints<br>– Position and orientation of patient |
| | | *Example 1 - Brain tumor segmentation: T1-weighted 3D acquisitions (1.0 x 1.0 x 1.0 mm^3), T1-weighted contrast-enhanced (gadolinium contrast) 3D acquisitions (1.0 x 1.0 x 1.0 mm^3), T2-weighted FLAIR 3D acquisitions (0.9 x 0.9 x 2.0 mm^3)*<br><br>*Example 2 - Instrument tracking: Only left frame recorded with da Vinci Si system*<br><br>*Example 3 - Modality classification in biomedical literature: Generally unknown* | – Radiation dose<br>– Radiopharmaceuticals<br>– Frequency of imaging<br>– Resolution<br>– Pixel spacing<br>– MRI: T1, T2, contrast enhanced, ...<br>– MRI: Repetition time (TR), echo time (TE)<br>– Field of view (FOV)<br>– Post-processing<br>– Employed dye(s)<br>– Use of filtering<br>– LM: phase contrast (PhC)<br>– CT: kEV<br>– Microscopy: Stainings<br>– Contrast agents |
| 28 | Operator(s) | Characteristics of operator(s) involved in the data acquisition process.<br><br>*Example 1 - Brain tumor segmentation: N/A*<br><br>*Example 2 - Instrument tracking: 2 male surgeons with more than 10 years of experience in laparoscopic surgery*<br><br>*Example 3 - Modality classification in biomedical literature: Unknown* | – Surgeon<br>– Engineer<br>– Nurses<br>– Robot<br>– Patient (e.g. with a smartphone app used for melanoma detection)<br>– OR<br>– Technician<br>– Medical trainee<br>– Biologist<br>– Radiologist<br>– Medical Physicist<br>– Radiographer<br>– Sonographer<br>– Unknown<br><br>Reported information may be:<br>– Number<br>– Function<br>– Names<br>– Skill level (e.g. measured in the number of years of experience) |
| *Validation data sets* | | | |
| 29 | Distribution of training and test cases* | Describes how training and test data were split and for what reason this division was chosen. This should include information (1) on why a specific proportion of training/test data was chosen, (2) why a certain total amount of cases was chosen and (3) why certain characteristics were chosen for the training/test set (e.g. class distribution according real-world distribution vs equal class distribution).<br><br>*Example 1 - Brain tumor segmentation: 80% training data and 20% test data according to common practice in machine learning.*<br><br>*Example 2 - Instrument tracking: All video sequences that are publicly available from site xy. Video sequences from institution x (50%) used for training. Video sequences from institution y (50%) used for testing.*<br><br>*Example 3 - Modality classification in biomedical literature: 60% training data and 40% test data according to common practice in the domain [ref]. Random assignment of images to training/test datasets.* | – Not applicable as no training data is provided<br>– Randomly distributed<br>– Balanced false and negative cases<br>– 80% training data, 20% test data as recommended by [ref] |
| 30 | Category of training data generation method* | Method for determining the desired algorithm output for the training data. Possible methods include manual image annotation, in silico ground truth generation and annotation by automatic methods, and no training data generated.<br><br>*Example 1 - Brain tumor segmentation: Hybrid: Initiation by algorithm and refinement/correction by expert physician*<br><br>*Example 2 - Instrument tracking: Crowdsourced annotations*<br><br>*Example 3 - Modality classification in biomedical literature: Manual annotation* | – Ground truth from simulation (exact)<br>– Reference from algorithm<br>– Reference from single human rater<br>– Reference from multiple human raters<br>– Hybrid: Initiation by algorithm, refinement by expert physician<br>– Reference derived from clinical practice (diagnosis/disease code etc.)<br>– Crowdsourced annotations |
| 31 | Number of training cases* | Number of cases that can be used for algorithm training and parameter optimization. A case encompasses all data that is processed to produce one result (e.g. one segmentation) as well as the corresponding reference result.<br><br>*Example 1 - Brain tumor segmentation: 400*<br><br>*Example 2 - Instrument tracking: 5 video sequences, each containing 100 annotated frames*<br><br>*Example 3 - Modality classification in biomedical literature: 6,000* | – No training data provided<br>– 100 images<br>– 100 raw endoscopic video sequences with a total of 1,000 fully annotated frames |
| 32 | Characteristics of training cases* | Additional information on the training cases describing their nature, such as the level of detail of the annotations (e.g. fully vs weakly annotated).<br><br>*Example 1 - Brain tumor segmentation: Pixel-level segmentation of the structures of interest and additional clinical information as described in context information.*<br><br>*Example 2 - Instrument tracking: Full segmentation of 100 frames (equally distributed) in each video sequence. No segmentation of the instruments in the remaining frames, but API information on instrument poses available for all frames.*<br><br>*Example 3 - Modality classification in biomedical literature: Full annotation - modality/image type per image* | – No training data provided<br>– Full annotation (pixel level)<br>– Weak annotation (image level): tumor volume, disease stage<br>– Mixed annotation: 1,000 fully annotated images, 100 weakly annotated images<br>– 100 endoscopic video images with 10 fully annotated training images |
| 33 | Annotation policy for training cases* | Instructions given to the annotators prior to training case annotation. This may include description of a training phase with the software.<br><br>*Example 1 - Brain tumor segmentation: The annotator was instructed to segment the edema using the T2 and FLAIR images. The enhancing tumor was subsequently to be segmented on the T1 contrast-enhanced modality. Finally, the necrotic core was to be outlined using the T1 and contrast-enhanced T1 image. The annotations were to be performed in axial slices. The undergraduate student received training on 5 cases (by the radiologist) to extract the weak labels (see parameter: context information).* | – Challenge-specific detailed instructions – e.g. should an annotation be performed along a tumor boundary or including a safety zone? Is it allowed to guess a boundary if not clearly visible?<br>– URL to annotation instructions<br>– What tissue would you resect?<br>– Where would you take a (small) biopsy? |
| | | *Example 2 - Instrument tracking: <URL to annotation instructions>* | |
| | | *Example 3 - Modality classification in biomedical literature: Instruction to label each cropped image with the modality that has (presumably) been used to acquire the image shown in the figure. List of potential labels: <list>. No further instructions, no training.* | |
| 34 | Annotator(s) of training cases* | Details on the subject(s)/algorithm(s) who/which annotated the training data.<br><br>*Example 1 - Brain tumor segmentation: Weak annotation (parameter context information) extracted from medical reports by undergraduate medical student; full image annotation performed by radiologist.*<br><br>*Example 2 - Instrument tracking: Crowdsourcing of image annotations on the platform Amazon Mechanical Turk according to the method in paper [ref]. Pose data (parameter context information) is automatically acquired by the robot*<br><br>*Example 3 - Modality classification in biomedical literature: Three PhD students* | − No training data provided<br>− Surgeon who has done >100 cases of a specific type of surgery<br>− Undergraduate physician (third year)<br>− Engineer who developed the software<br>− Physician with no prior experience in usage of the software<br>− Crowd<br>− Algorithm xy |
| 35 | Annotation aggregation method(s) for training cases* | Method(s) used to merge multiple annotations for one case.<br><br>*Example 1 - Brain tumor segmentation: (only one observer)*<br><br>*Example 2 - Instrument tracking: According to [ref]*<br><br>*Example 3 - Modality classification in biomedical literature: Majority vote* | − No aggregation<br>− Simultaneous Truth and Performance Level Estimation (STAPLE)<br>− Majority vote<br>− An additional annotator resolves conflicts<br>− Average<br>− Selective and Iterative Method for Performance Level Estimation (SIMPLE)<br>− Level-set based approach maximizing the a posteriori probability (LSML)<br>− Strict combination (positive if and only if all annotators agree)<br>− No training data is required |
| 36 | Category of reference generation method* | Method for determining the reference (i.e. the desired algorithm result, also referred to as gold standard) which is used for assessing the participants' algorithms. Possible methods include manual image annotation, in silico ground truth generation and annotation by automatic methods.<br><br>*Example 1 - Brain tumor segmentation: Manual annotation*<br><br>*Example 2 - Instrument tracking: Crowdsourced annotations*<br><br>*Example 3 - Modality classification in biomedical literature: Manual annotation by multiple observers* | − Ground truth from simulation (exact)<br>− Reference from algorithm<br>− Reference from single human rater<br>− Reference from multiple human raters<br>− Acquired through previously validated methods according to [ref]<br>− Reference derived from clinical practice (diagnosis/disease code etc.)<br>− Crowdsourced annotations<br>− Hybrid methods (e.g. initiation by algorithm, refinement by expert physician) |
| 37 | Number of test cases* | Number of cases used to assess the performance of an algorithm. A case encompasses all data that is processed to produce one result as well as the corresponding reference result (typically not provided to the participants). | − 100 images<br>− 100 raw endoscopic video sequences with a total of 1,000 fully annotated frames |
| | | *Example 1 - Brain tumor segmentation: 100* | |
| | | *Example 2 - Instrument tracking: 5 video sequences, each containing 100 annotated frames.* | |
| | | *Example 3 - Modality classification in biomedical literature: 4,000* | |
| 38 | Characteristics of test cases* | Additional information on the test cases describing their nature, such as the level of detail of the annotations (e.g. fully vs weakly annotated). *Example 1 - Brain tumor segmentation: Pixel-level segmentation of the structures of interest and additional clinical information as described in context information* *Example 2 - Instrument tracking: Full segmentation of 100 frames (equally distributed) in each video sequence. No segmentation of the instruments in the remaining frames, but API information on instrument poses available for all frames* *Example 3 - Modality classification in biomedical literature: Full annotation - modality/image type per image* | – Full annotation (pixel level)<br>– Weak annotation (image level): tumor volume, disease stage<br>– Mixed annotation: 1,000 fully annotated images, 100 weakly annotated images<br>– 100 endoscopic video images with 10 fully annotated test images |
| 39 | Annotation policy for test cases* | Instructions given to the annotators prior to test case annotation. This may include description of a training phase with the software. *Example 1 - Brain tumor segmentation: The annotator was instructed to segment the edema using the T2 and FLAIR images. The enhancing tumor was subsequently to be segmented on the T1 contrast-enhanced modality. Finally, the necrotic core was to be outlined using the T1 and contrast-enhanced T1 image. The annotations were to be performed in axial slices. The undergraduate student had received training on extracting the weak labels when annotating the training images.* *Example 2 - Instrument tracking: <URL to annotation instructions>* *Example 3 - Modality classification in biomedical literature: Instruction to label each cropped image with the modality that has (presumably) been used to acquire the image shown in the figure. List of potential labels: {...}. After two observers have annotated the images independently, the cases where the annotators disagreed will be automatically retrieved. These should be shown to the third observer for resolving conflicts. Training on the training cases.* | – Challenge-specific detailed instructions – e.g. should an annotation be performed along a tumor boundary or including a safety zone? Is it allowed to guess a boundary if not clearly visible?<br>– URL to annotation instructions<br>– What tissue would you resect?<br>– Where would you take a (small) biopsy? |
| 40 | Annotator(s) of test cases* | Details on the subject(s)/algorithm(s) who/which annotated the test data. *Example 1 - Brain tumor segmentation: Radiologist with 5 years of experience* *Example 2 - Instrument tracking: Crowdsourcing on the platform Amazon Mechanical Turk* | – Surgeon who has done >100 cases of a specific type of surgery<br>– Undergraduate physician (third year)<br>– Engineer who developed the software<br>– Physician with no prior experience in usage of the software<br>– Crowd |
| | | *according to the method [ref]*<br><br>*Example 3 - Modality classification in biomedical literature: Two PhD students and a radiologist* | − Algorithm xy |
| 41 | Annotation aggregation method(s) for test cases* | Method(s) used to merge multiple annotations for one case (if any).<br><br>*Example 1 - Brain tumor segmentation: No merging*<br><br>*Example 2 - Instrument tracking: According to [ref]*<br><br>*Example 3 - Modality classification in biomedical literature: Expert resolves conflicts* | − No aggregation (ranking provided for each annotator)<br>− STAPLE<br>− Majority vote<br>− An additional annotator resolves conflicts<br>− Average<br>− SIMPLE<br>− LSML<br>− Strict combination (positive if and only if all annotators agree) |
| 42 | Data pre-processing method(s) | Methods used for pre-processing the raw data before it is provided to the participants.<br><br>*Example 1 - Brain tumor segmentation: Registration of different contrasts using algorithm x and denoising using algorithm y. Both performed in framework z. Resampling to common coordinate system with spatial resolution 1x1x1mm^3; skull stripping according to [ref], bias field correction according to [ref].*<br><br>*Example 2 - Instrument tracking: Irrelevant scene removal using algorithm xy [ref] with parameters z.*<br><br>*Example 3 - Modality classification in biomedical literature: Cropping to figures; if compound figures are included, then these are separated manually into the subfigures using the software xy.* | − No pre-processing steps<br>− Registration with a particular method<br>− Segmentation with a particular method<br>− Resampling of raw data<br>− Re-orientation<br>− Normalization<br>− Data cleaning<br>− Instance selection<br>− Feature extraction<br>− Feature selection<br>− Video anonymization<br>− Bias correction<br>− Intensity standardization<br>− White balancing<br>− Smoothing<br>− Skull stripping<br>− Histogram Matching<br>− Background subtraction<br>− Uneven background intensity correction<br>− Image enhancement (contrast/brightness change, histogram equalization)<br>− Data format conversion (DICOM to NIFTI)<br>− Journals: cropping, grey level reduction |
| 43 | Potential sources of reference errors | Most relevant possible error sources related to the estimation of the reference. This may include errors related to the image acquisition method, user errors and errors resulting from the pre-processing method applied. It may be quantified by inter- and intra-observer variability, for example. This information must be provided for the test cases and may be provided for the training cases.<br><br>*Example 1 - Brain tumor segmentation: Tissue classes (tumor necrosis, enhancing tumor and edema) difficult to distinguish. Previous studies suggest inter-rater disagreement of xy [ref].*<br><br>*Example 2 - Instrument tracking: The provided API robot pose data has an estimated accuracy of x according to additional experiments performed with the calibration phantom described in [ref]. The accuracy of crowdsourced segmentations of medical instruments with the applied method has been estimated to be around z [ref].*<br><br>*Example 3 - Modality classification in biomedical literature: Previous experiments [ref] suggest* | Sources of error:<br>− Partial volume effects<br>− Errors resulting from surface generation<br>− User errors<br>− Organ deformation<br>− Distortion of electromagnetic field<br>− Noise<br>− Imaging system aberrations (deteriorated point spread function resulting in blurring)<br>− Imaging system artifacts (dust, motion, spikes, etc.)<br>− Interlacing<br><br>Quantification:<br>− Inter-/ intra-observer variability<br>− Confidence intervals<br>− Kappa statistics<br>− Correlation coefficients<br>− Signal-to-noise ratio<br>− Resolution<br>− Bland-Altman Plots |

| | | *that the main source of error is related to ambiguity in the images, as there is an extremely large variability and sometimes the defined image types are hard to fit into the existing hierarchy of a small number of image types; other mistakes can be linked to annotating quickly and not looking at sometimes fine differences. Inter-observer variability in a separate experiment was x%.* | |
|---|---|---|---|
| *Assessment method* | | | |
| 44 | Metric(s)* | Function(s) to assess a property of an algorithm. These functions should reflect the validation objective (see parameter *assessment aim(s)*).<br><br>*Example 1 - Brain tumor segmentation: 95% HD and precision applied separately to necrosis, enhancing tumor and edema.*<br><br>*Example 2 - Instrument tracking: Runtime per frame on an xy device; Percentage of frames with DSC below threshold t.*<br><br>*Example 3 - Modality classification in biomedical literature: Accuracy* | − Hausdorff distance (HD)<br>− Dice similarity coefficient (DSC)<br>− Jaccard index<br>− Computation time<br>− Recall<br>− Precision<br>− Area under curve (AUC)<br>− Root mean square error (RMSE)<br>− Absolute volume difference<br>− True positive rate<br>− Computational complexity<br>− Average symmetric surface distance (ASSD)<br>− F1-Score<br>− Specificity<br>− Intraclass correlation coefficient<br>− Concordance index<br>− MAP |
| 45 | Justification of metric(s)* | Justification why the metric(s) was/were chosen, preferably with reference to the clinical application.<br><br>*Example 1 - Brain tumor segmentation: 95% Hausdorff Distance as opposed to standard HD: Try to avoid that outliers have too much weight. All other metrics are commonly used in segmentation assessment (cf. ref. xy).*<br><br>*Example 2 - Instrument tracking: Thresholded DSC according to best practice recommendations [ref]. Computation time as clinical application requires video rate performance.*<br><br>*Example 3 - Modality classification in biomedical literature: Accuracy is the most common method as easy to interpret; geometric mean assures that all classes are well classified and not only the majority classes.* | − According to best practice recommendations [ref]<br>− According to paper [ref] |
| 46 | Rank computation method* | Method used to compute a rank for all participants based on the generated results on the test data. It may include methods for aggregating over all test cases and/or for determining a final rank from multiple single metric-based ranks. It also includes the ranking order for tied positions.<br><br>*Example 1 - Brain tumor segmentation: For each participant pi and each test case cj: Compute the metric values for the 95% Hausdorff distance and precision. For each participant pi and each test case cj, determine the rank corresponding to both metrics (i.e. R(precision, pi, cj): descending order for precision, R(HD, pi, cj) ascending for 95% HD).* | Example 1:<br>0. Initialization: For each participant pi and each test case cj: compute metric values M1(pi,cj) and M2(pi,cj) for metrics M1 and M2.<br>1. Metric-based aggregation: For each participant pi compute the median over all cases cj for each metric M1(pi) and M2(pi).<br>2. For each participant pi, compute the sum over the two metrics as M1(pi) + M2(pi).<br>3. Build rank for each participant by sorting the values M1(pi) + M2(pi) for each participant.<br><br>Example 2:<br>0. Initialization: For each participant pi and |
| | | *For each participant pi and each test case cj, compute the average rank R(pi, cj) over both metric ranks. Finally, compute the average over all case-specific ranks to get one final rank for each participant pi.*<br><br>*Example 2 - Instrument tracking: For each participant pi compute the average metric value for the thresholded DSC. Build the rank for each participant by sorting the accumulated metric values. In case of tied positions, perform the ranking according to computation times.*<br><br>*Example 3 - Modality classification for retrieval tasks: Ranking performed according to [ref].* | each test case cj: compute metric values M1(pi,cj), M2(pi,cj) and M3(pi,cj) for metrics M1-M3.<br>1. Case-based aggregration: For each partipant pi and each case cj, determine the performance score sj on case cj: sij := 1/3 (M1(pi,cj) + M2(pi,cj) + M3(pi,cj)).<br>2. For each partipant pi and each case cj, determine the rank R(pi,cj) for case cj according to score sj.<br>3. Compute the average over all case-specific ranks for each participant pi si := 1/N*Sum_j (R(pi,cj)) to obtain the final rank. |
| 47 | Interaction level handling* | Method(s) used to handle any diversity in the level of user interaction when generating the performance ranking.<br><br>*Example 1 - Brain tumor segmentation: Weighting function (automatic methods are ranked higher)*<br><br>*Example 2 - Instrument tracking: Only automatic algorithms are allowed.*<br><br>*Example 3 - Modality classification in biomedical literature: Only automatic methods are allowed.* | − Indication in ranking<br>− Separate ranking for fully-automatic methods<br>− Only automatic methods allowed |
| 48 | Missing data handling* | Methods used to manage submissions with missing results on test cases.<br><br>*Example 1 - Brain tumor segmentation: In case of missing data for participant pi and case cj, the case-based ranks for all metrics m R(m, pi, cj) are set to the maximum.*<br><br>*Example 2 - Instrument tracking: Only complete submissions are evaluated.*<br><br>*Example 3 - Modality classification in biomedical literature: Missing results are considered to be incorrectly classified* | − Missing data not allowed (incomplete submissions not evaluated)<br>− Missing data ignored<br>− Missing data handled as in [ref] |
| 49 | Uncertainty handling* | Method(s) used to make uncertainties in ranking explicit.<br><br>*Example 1 - Brain tumor segmentation: Test the sensitivity of the ranking with bootstrapping according to [ref].*<br><br>*Example 2 - Instrument tracking: Test the sensitivity of a ranking by leaving out different amounts of test data as described in [ref].*<br><br>*Example 3 - Modality classification in biomedical literature: None* | Test sensitivity of the ranking by<br>− Leaving out test data<br>− Bootstrapping approaches<br>− Changes in rank computation details<br>− Changes in reference annotation |
| 50 | Statistical test(s)* | Statistical test(s) used to compare the results of challenge participants.<br><br>*Example 1 - Brain tumor segmentation: T-test used to test the stability of the first three ranks as described in [ref].*<br><br>*Example 2 - Instrument tracking: U-test used to test statistically significant differences between* | Quantities on which the hypothesis is taken:<br>− Stability of the ranking<br>− See whether the best results have statistically significant differences<br><br>Tests:<br>− Wilcoxon-Mann-Whitney test<br>− t-test (paired, unpaired, one-sided, two-sided) |
| | | *the participants as described in [ref].*<br>*Example 3 - Modality classification in biomedical literature: No statistical tests.* | – Saphiro-Wilk test |
| **Challenge outcome** | | | |
| 51 | Information on participants | Information on participating teams including affiliation and specifics of competing algorithm, preferably with reference to a document. | Should include:<br>– Acronym<br>– Affiliation<br>– Contact person<br>– Team members<br>– Method description including parameter instantiation<br>– Submission/attempt number<br>– Relevant reference publication |
| 52 | Results | Values of all metrics and rankings including the number of submissions for each participant.<br><br>*Example 1 - Brain tumor segmentation: Not yet available*<br><br>*Example 2 - Instrument tracking: Matrix with colums = participants; rows = (case j, metric k); in addition two rankings for the two metrics and the final ranking.*<br><br>*Example 3 - Modality classification in biomedical literature: Two rankings corresponding to the two metrics and the associated aggregated values.* | – Not yet available<br>– Link to URL<br>– Matrix with results for each participant, each metric and each test case plus ranking |
| 53 | Publication | Evaluating and summarizing information about the challenge or the workshop published in a scientific journal or similar literature, preferably with DOI.<br><br>*Example 1 - Brain tumor segmentation: IEEE Transactions on Medical Imaging: <link to pdf>*<br><br>*Example 2 - Instrument tracking: arXiv publication: <URL>*<br><br>*Example 3 - Modality classification in biomedical literature: Publication in CEUR: <DOI>* | – No publication<br>– DOI<br>– Link to document<br>– Full citation<br>– arXiv ID |

# Supplement 2

**Questionnaire: Towards next-generation biomedical challenges**

The questions of the online questionnaire were related to the following categories:

General information

- What is your position?
- Which country do you live in?
- What is your primary background?

Background with respect to challenge participation

- Have you taken part in a challenge to date? If so, how many challenges?
- How much effort do you put into a challenge participation?
- Before the challenge: What was your motivation to participate in a challenge?
- During the challenge: What did you struggle the most with before submitting your results?
- After submitting your results: Did you encounter problems interpreting your challenge rank?
- Have you ever registered as a participant of a challenge for which you did not submit results? If so, how many times? What were the main reasons for not submitting results?

Background with respect to challenge organization

- Have you ever taken part in the organization of a challenge? If so, how many challenges?
- Have you ever participated in your own challenge?
- In percent, please estimate the time the challenge organizers (as a group) put in the
    i. design of the challenge
    ii. preparation of the challenge
    iii. processing of submissions
    iv. on-site execution of the challenge
- Please estimate how many hours the organizing team spent in total on preparing and executing the challenge.
- Did you struggle with any of the following problems during designing your challenge?
    i. Choosing the metrics
    ii. Finding data sets
    iii. Creating the reference data
    iv. Deciding on how to create the challenge ranking
    v. None of them
    vi. Other
- Did you encounter problems during preparing your challenge? Which ones?
- Did you encounter problems during the processing of the submissions of your challenge? Which ones?
- Did you encounter problems during the on-site execution of your challenge (if any)? Which ones?
- Can you think of aspects that you would like to improve in the future?

Issues related to challenge design and organization

- What do you consider issues related to the data of biomedical challenges?
- What do you consider issues related to the annotation (reference data) of biomedical challenges?
- What do you consider issues related to the evaluation of biomedical challenges?
- What do you consider issues related to the documentation of biomedical challenges?

General view on challenges

- Should challenge organizers provide pre-evaluation results?
- Should challenge organizers (and group members) be allowed to participate in their own challenge?
- Please explain under which conditions challenge organizers (and group members) should be allowed to participate in their own challenge.
- How serious do you rate the fact that algorithms can be tuned to the challenge data?
- Do you generally think that challenge rankings reflect algorithm performances well?
- Do you think the design of current biomedical challenges should be improved in general?

Open issues and recommendations

- What recommendations do you have for the improvement of biomedical challenges?
- What are open research issues with respect to biomedical challenge design?
- Would you appreciate best practice guidelines for biomedical challenges?
- Should challenges organized in the scope of big conferences (e.g. MICCAI) undergo more quality control?
- What actions could the research community (i.e. MICCAI society) undertake to improve challenge quality in general?

# Supplement 3

**Results of Questionnaire: Towards next-generation biomedical challenges**

For the following analysis, only complete questionnaires (n = 117) and questionnaires with > 50% answers  (n = 12) were considered. The majority of the participants were professors (30%), PhD students (23%) and postdoctoral researchers (14%) and had a background in engineering, maths, physics or computing (94%). 31% of the participants had already organized a challenge and 63% had taken part in at least one challenge. 92% of all participants agreed that biomedical challenge design should be generally improved. The following problems were identified for the categories data, annotation, evaluation and documentation. For all four categories, we report the most commonly reported problems ordered by frequency of reporting.

**Data**

*Representativeness (33%):* Most concerns were related to the representativeness of the data. Criticism was targeted mainly at the representativeness of the training and test sets, balance of the data (e.g. between positives and negatives), selection bias, realism of the data (e.g. with respect to noise and artefacts) and the typically small number of centers/vendors/devices involved. In fact, our analysis revealed that the median number of institutes involved in a challenge was 1 (IQR: (1, 1)) and only 17% of all tasks conducted up to 2016 were based on multi-center data.

One critical consequence of the generally small data sets is that challenge participants tend to complement available training data with their own data sets which makes it impossible to distinguish the effect of an algorithm from the effect of the training data.

*Data acquisition (17%):* Participants agreed that the data acquisition itself is one of the main barriers for challenge organizers, especially due to legal barriers and high costs. In fact, 22% of challenge organizers encountered problems acquiring the data sets for their challenge. One of the main reasons for participation in a challenge was the access to validation data (30% of challenge participants).

Further problems mentioned include

- Heterogeneity (8%), e.g. due to the lack of acquisition standards
- Infrastructure issues (8%), mainly due to inconsistencies in formats and the lack data management and exploration tools
- Accessibility of the data (8%), mainly after the challenge
- Lack of documentation (8%), especially about the image acquisition process
- Data quality (6%), in general and
- Overfitting/Cheating/Tuning (4%).

**Annotation**

*Quality of reference data (33%):* Major concerns were related to errors in the reference data. The annotations were regarded as subjective and/or biased, e.g. because only single observers annotated the data in many cases or automatic tools were used either for the annotations themselves or for the initialization in semi-automatic tools. The lack of quality control in this step was regarded particularly critically for many challenges.

*Method for reference generation (16%):* A related issue is the method chosen for reference annotation. For example, studies have shown (e.g. [Lampert et al. 2016]) that reference annotations may vary significantly even across medical experts. This issue can potentially be compensated to some extent by merging reference annotations from multiple experts but this was only done (reported) in 73% of all tasks. It should also be noted that 27% of all challenge organizers encountered problems when generating the reference data for their challenge.

*Transparency (15%):* Lack of transparency was another major issue raised in the context of data annotation. In particular, it was requested to make raw annotations available, report on inter- and intra-observer variability and document how the final reference annotation was generated.

*Resources (14%):* Another issue raised were the resources required for providing high-quality reference data. The annotation was not only considered particularly challenging but also logistically hard due to improper tools.

*Lack of standards (10%):* The lack of guidelines for annotating and merging annotations was heavily criticized. Similarly, the lack of standard data formats was regarded critical.

**Evaluation**

*Choice of metric (20%):* The metrics applied in current challenges were generally criticized. For example, it was stated that runtime/computational complexity are rarely considered. The metrics are also often not well linked to the clinical context. Finally, optimal metric aggregation is a major issue to be addressed, especially in the case of missing data. On the other hand, participants agreed that finding the right metric(s) for a given task is highly challenging. In fact, 23% of all challenge organizers struggled with the choice of metric(s).

*Lack of standards (19%):* A major point of criticism was related to the lack of standards with respect to metrics and evaluation frameworks. For example, even the presumably same metrics are sometimes named or applied differently.

*Transparency (12%):* Missing documentation with respect to the evaluation process was criticized. Concerns were raised regarding the fact that the evaluation criteria are not transparent before the submission of data, potentially allowing organizers to influence the final ranking.

Further difficulties raised include

- Lack of infrastructure/tools (7%) such as public tools for evaluation
- Method for determining ranking (7%), for instance, in the case of missing values
- Lack of quality control (7%), e.g. related to metric implementation
- Too much focus on ranking (5%), especially when considering the sensitivity of the ranks
- Lack of uncertainty handling (4%), especially when considering inter-observer variability as well as
- Lifetime and dynamics (3%) of challenges.

**Documentation**

*Completeness and transparency (47%):* Participants agreed that the documentation should be as comprehensive as possible, which is currently not the case. This holds true not only for the reporting of challenge design and results but also for the methods themselves. For example, a challenge ranking should reflect the quality of a method in the context of a given task. Unfortunately, however, method performance may depend crucially on its parameters. Given that only 4% of challenge participants stated that they typically apply their algorithm "as is" to new challenge data, and more than 80% of participants tune their methods to a given task, it comes at a surprise that almost no attention is given to the parameters applied by the algorithms assessed.

*Publication of results (13%):* Issues with respect to the publication of results further include the delay between submission deadline and paper publication as well as discrepancies between the challenge website and the corresponding publication.

*Lifetime and dynamics (10%):* Another problem raised was the typically dynamically changing content of challenge websites, which makes proper referencing hard. An open research question is further the optimal lifetime of a challenge considering problems with overfitting, for example.

Further problems related to the documentation were

– Lack of open source code (9%) corresponding to participating algorithms and the evaluation
– Lack of standards for structured reporting (7%), such as common ontologies
– Accessibility of information (5%) especially after the challenge as well as the
– Lack of acknowledgement/citation of all people involved (3%).

## Author contributions

## Acknowledgements

We thank all organizers of the 2015 segmentation challenges who are not co-authoring this paper, in particular Cheng-Ta Huang (National Taiwan University of Science and Technology, Taiwan), Chung-Hsing Li (Tri-Service General Hospital, Taiwan), Sheng-Wei Chang (Tri-Service General Hospital, Taiwan), Svitlana Zinger (Eindhoven University of Technology, The Netherlands), Erik Schoon (Catharina Hospital Eindhoven, The Netherlands), Peter de With (Eindhoven University of Technology, The Netherlands), Gustavo Carneiro and Zhi Lu (University of Adelaide, Australia), Jing Wu (Medical University of Vienna, Austria), Ana-Maria Philip (Medical University of Vienna, Austria), Bianca S. Gerendas (Medical University of Vienna, Austria), Sebastian M. Waldstein (Medical University of Vienna, Austria), Ursula Schmidt-Erfurth (Medical University of Vienna, Austria), all involved readers of the OPTIMA team and the VRC (Vienna Reading Center, Austria), Patrik Raudaschl (Institute for Biomedical Image Analysis, UMIT, Austria), Karl Fritscher (Institute for Biomedical Image Analysis, UMIT, Austria), Paolo Zaffino (Magna Graecia University of Catanzaro, Italy), Maria Francesca Spadea (Magna Graecia University of Catanzaro, Italy), Dzung L. Pham (CNRM, The Henry M. Jackson Foundation for the Advancement of Military Medicine, USA), Jerry L. Prince (Johns Hopkins University, USA), Jean-Christophe Houde (Université de Sherbrooke, Canada), Emmanuel Caruyer (CNRS Paris, France), Alessandro Daducci (École Polytechnique Fédérale de Lausanne, Switzerland), Tim Dyrby (Danish Research Centre for Magnetic Resonance, Denmark), Bram Stieltjes (University Hospital Basel, Switzerland), Maxime Descoteaux (Université de Sherbrooke, Canada), Orcun Goksel (ETH Zürich, Switzerland), Antonio Foncubierta-Rodríguez (ETH Zürich, Switzerland), Oscar Alfonso Jiménez del Toro (HES-SO Valais-Wallis, Switzerland), Georg Langs (Medical University of Vienna, Austria), Ivan Eggel (HES-SO Valais-Wallis, Switzerland), Katharina Gruenberg (Radiologisches Zentrum Wiesloch, Germany), Marianne Winterstein (Universitätsklinikum Heidelberg, Germany), Markus Holzer (contextflow GmbH, Austria), Markus Krenn (contextflow GmbH, Austria), Georgios Kontokotsios (EBCONT enterprise technologies GmbH, Austria), Sokratis Metallidis (EBCONT enterprise technologies GmbH, Austria), Roger Schaer (HES-SO Valais-Wallis, Switzerland), András Jakab (Neuroscience Center Zürich, Switzerland), Tomàs Salas Fernandez (Agency for Health Quality and Assessment of Catalonia, Spain), Sebastian Bodenstedt (NCT Dresden, Germany), Martin Wagner (University Hospital Heidelberg, Germany), Hannes Kenngott (University Hospital Heidelberg, Germany), Max Allan (Intuitive Surgical, Inc., USA), Mauricio Reyes (Bern University, Switzerland), Keyvan Farahani (NIH, USA), Jayashree Kalpathy-Cramer (Harvard MGH, USA), Dongjin Kwon (University of Pennsylvania, USA), Heinz Handels (Universität zu Lübeck, Germany), Matthias Liebrand (Universitätsklinikum Schleswig-Holstein, Germany), Ulrike Krämer (Universitätsklinikum Schleswig-Holstein, Germany), Shuo Li (University of Western Ontario, Canada), Stephen M. Damon (Vanderbilt University, USA).

We further thank Angelika Laha, Diana Mindroc-Filimon, Bünyamin Pekdemir and Jenshika Yoganathan (DKFZ, Germany) for helping with the comprehensive challenge capturing. Many thanks also go to Janina Dunning and Stefanie Strzysch (DKFZ, Germany) for their support of the project.

We thank all participants of the international questionnaire, in particular those who filled out > 50% of the form including Daniel Alexander (University College London, UK), Susan Astley (University of Manchester, UK), Peter Bandi (Radboud University Medical Center, The Netherlands), Floris F. Berendsen (Leiden University Medical Center, The Netherlands), Jorge Bernal (Universitat Autonoma de Barcelona, Spain), Zijian Bian (Northeastern University, PRC), Andrew P. Bradley (The University of Queensland, Australia), Paul A. Bromiley (University of Manchester, UK), Esther Bron (Erasmus Medical Center, The Netherlands), Philippe Cattin (University of Basel, Switzerland), Eric Chang (Microsoft Research, PRC), Christos Chatzichristos (National and Kapodistrian University Athens, Greece), Stephane Chauvie (Santa Croce e Carle Hospital, Italy), Xingqiang Chen (Xiamen University, PRC), Veronika Cheplygina (Eindhoven University of Technology, The Netherlands), Guy Cloutier (University of Montreal, Canada), Marleen De Bruijne (Erasmus Medical Center Rotterdam, The Netherlands, and University of Copenhagen, Denmark), Maxime Descoteaux (Université de Sherbrooke, Canada), Fabien Despinoy (LTSI-Inserm U1099, Rennes, France), Thijs Dhollander (The Florey Institute of Neuroscience and Mental Health, Australia), Xiaofei Du (University College London, UK), Sara El Hadji (Politecnico di Milano, Italy), Ahmed El Kaffas (Stanford University, USA), Andrey Fedorov (Brigham and Women's Hospital, USA), Simon Fristed Eskildsen (Aarhus University, Denmark), Babak Ehteshami Bejnordi (Radboud University Medical Center, The Netherlands), Luc Florack (Eindhoven University of Technology, The Netherlands), Yaozong Gao (Apple, USA), Bernard Gibaud (INSERM, Rennes, France), Paul-Gilloteaux (CNRS, Nantes, France), Michael Götz (DKFZ, Germany), Horst Hahn (Fraunhofer MEVIS & Jacobs University Bremen, Germany), Alexander Hammers (King's College London, UK), Yuankai Huo (Vanderbilt University, USA), Allan Hanbury (Technical University Wien, Austria), Luis C. Garcia-Peraza Herrera (University College London, UK), Oscar Jimenez (HES-SO, Switzerland), Leo Joskowicz (The Hebrew University of Jerusalem, Israel), Bernhard Kainz (Imperial College London, UK), Sjoerd Kerkstra (DIAG - Radboud University Medical Center, The Netherlands), Stefan Klein (Erasmus MC, The Netherlands), Michal Kozubek (Masaryk University, Czech Republic), Walter G. Kropatsch (Technical University Wien, Austria), Pablo Lamata (King's College London, UK), Michele Larobina (Italian National Research Council, Italy), Gaby Martins (Instituto Gulbenkian de Ciencia, Portugal), Keno März (DKFZ, Germany), Matthew McCormick (Kitware, Inc., USA), Stephen McKenna (University of Dundee, UK), Karol Miller (UWA, Australia), Erika Molteni (University College London, UK), Cosmin Adrian Morariu (University of Duisburg-Essen, Germany), Henning Müller (HES-SO, Switzerland), Sérgio Pereira (University of Minho, Portugal), Ingerid Reinertsen (SINTEF, Norway), Mauricio Reyes (University of Bern, Switzerland), Constantino Carlos Reyes-Aldasoro (University of London, UK), Gerard Ridgway (University of Oxford, UK), Robert Rohling (UBC, Canada), James Ross (Brigham and Women's Hospital, USA), Danny Ruijters (Philips Healthcare, The Netherlands), Olivier Salvado (CSIRO, Australia), Gerard Sanroma (Universitat Pompeu Fabra, Spain), Ayushi Sinha (Johns Hopkins University, USA), Chetan L. Srinidhi (National Institute of Technology Karnataka, Surathkal, India), Iain Styles (University of Birmingham, UK), Paul Summers (IEO, Italy), Agnieszka Szczotka (University College London, UK), Raphael Sznitman (University of Bern, Switzerland), Sabina Tangaro (Istituto Nazionale di Fisica Nucleare, Italy), Lennart Tautz (Fraunhofer MEVIS, Germany), Sotirios Tsaftaris (The University of Edinburgh, UK), Fons van der Sommen (Eindhoven University of Technology, The Netherlands), Koen Van Leemput (Massachusetts General Hospital,

USA), Theo van Walsum (Erasmus Medical Center, The Netherlands), Ching-Wei Wang (National Taiwan University of Science and Technology, Taiwan), Li Wang (UNC, USA), Michael Wels (Siemens Healthcare GmbH, Germany), Rene Werner (University Medical Center Hamburg-Eppendorf, Germany), Thomas Wollmann (University of Heidelberg, BIOQUANT, IPMB, and DKFZ, Germany), Alistair Young (University of Auckland, New Zealand), Lining Zhang (University College London, UK), Dženan Zukić (Kitware, Inc., USA), Maria A. Zuluaga (Amadeus, France).

We would further like to acknowledge support from the European Research Council (ERC) (ERC starting grant COMBIOSCOPY under the New Horizon Framework Programme grant agreement ERC-2015-StG-37960 as well as Seventh Framework Programme (FP7/2007-2013) under grant agreement n° 318068 (VISCERAL)), the German Research Foundation (DFG) (grant MA 6340/10-1 and grant MA 6340/12-1), the Ministry of Science and Technology, Taiwan (MOST 106-3114-8-011-002, 106-2622-8-011-001-TE2, 105-2221-E-011-121-MY2), the US National Institute of Health (NIH) (grants R01-NS070906, RG-1507-05243 and R01-EB017230 (NIBIB)), the Australian Research Council (DP140102794 and FT110100623), the Swiss National Science Foundation (grant 205321_157207), the Czech Science Foundation (grant P302/12/G157), the Czech Ministry of Education, Youth and Sports (grant LTC17016 in the frame of EU COST NEUBIAS project), the Engineering and Physical Sciences Research Council (EPSRC) (MedIAN UK Network (EP/N026993/1) and EP/P012841/1), the Wellcome Trust (NS/A000050/1), the Canadian Natural Science and Engineering Research Council (RGPIN-2015-05471), the UK Medical Research Council (MR/P015476/1), and the Heidelberg Collaboratory for Image Processing (HCI) including matching funds from the industry partners of the HCI.